\documentclass[10pt,twocolumn,letterpaper]{article}

\usepackage[pagenumbers]{iccv} 

\definecolor{iccvblue}{rgb}{0.21,0.49,0.74}
\usepackage[pagebackref,breaklinks,colorlinks,allcolors=iccvblue]{hyperref}

\usepackage{graphicx}
\newsavebox\CBox
\def\textBF#1{\sbox\CBox{#1}\resizebox{\wd\CBox}{\ht\CBox}{\textbf{#1}}}
\usepackage{numprint}
\usepackage{multirow}

\newcolumntype{P}[1]{>{\centering\arraybackslash}p{#1}}
\usepackage{makecell}
\usepackage{hhline}
\usepackage{bm}

\usepackage{amsmath}

\DeclareMathOperator*{\argmin}{arg\,min}

\usepackage{lipsum}

\title{OMNI-DC: Highly Robust Depth Completion with Multiresolution Depth Integration}
\newcommand{\methodname}{OMNI-DC}

\author{Yiming Zuo, Willow Yang, Zeyu Ma, Jia Deng \\
Department of Computer Science, Princeton University\\
{\tt\small \{zuoym,willowliuyang,zeyum,jiadeng\}@princeton.edu}}

\begin{document}

\twocolumn[{
\renewcommand\twocolumn[1][]{#1}
\maketitle
\vspace{-11mm}
\begin{center}
    \centering
    \captionsetup{type=figure}
    \includegraphics[width=\textwidth]{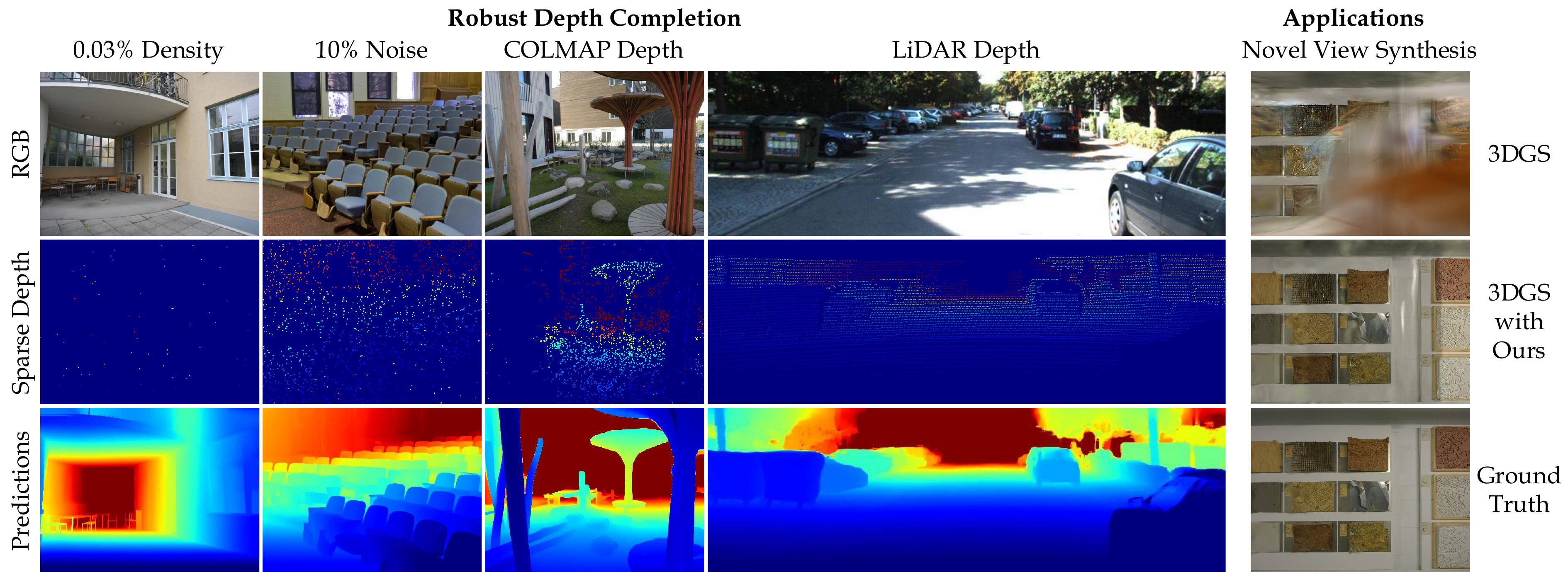}
    \vspace{-6mm}
    \captionof{figure}{Given an RGB image and a sparse depth map, \methodname\ generates high-quality dense depth maps on different types of scenes (indoors/outdoors/urban) with a variety of sparse depth patterns, in a \textit{zero-shot} manner: none of these datasets is seen during training. The dense depth maps can be used to train 3DGS~\cite{3dgs}, greatly improving the rendering quality for novel view synthesis.}
    \vspace{-2mm}
\end{center}
}]

\begin{abstract}
  Depth completion (DC) aims to predict a dense depth map from an RGB image and a sparse depth map. Existing DC methods generalize poorly to new datasets or unseen sparse depth patterns, limiting their real-world applications. We propose \methodname, a highly robust DC model that generalizes well zero-shot to various datasets. The key design is a novel Multi-resolution Depth Integrator, allowing our model to deal with very sparse depth inputs. We also introduce a novel Laplacian loss to model the ambiguity in the training process. Moreover, we train \methodname\ on a mixture of high-quality datasets with a scale normalization technique and synthetic depth patterns. Extensive experiments on 7 datasets show consistent improvements over baselines, reducing errors by as much as 43\%. Codes and checkpoints are available at \url{https://github.com/princeton-vl/OMNI-DC}.
\end{abstract}

\vspace{-5mm}
\section{Introduction}
\label{sec:intro}

Depth completion (DC) is the task of predicting a dense depth map from an RGB image and a sparse depth map. Dense and accurate depth maps are crucial for many downstream applications, such as robotics and 3D reconstruction, where sparse depth hints are often available. DC has been successfully applied to various tasks, including autonomous driving~\cite{DC_app_1,DC_app_2}, 3D reconstruction~\cite{DC_app_3d_recon}, and novel view synthesis~\cite{dense_depth_prior_nerf,dn-splatter}, where sparse depth data come from either active sensors such as LiDAR~\cite{DC_app_1,DC_app_2,DC_app_3d_recon}, or multiview matching (\eg, Structure from Motion, SfM)~\cite{dense_depth_prior_nerf,dn-splatter}.

In recent years, various methods for DC have been proposed~\cite{ognidc,BP-Net,DFU,completionformer,LRRU,DySPN,NLSPN}. While these methods achieve impressive accuracy on a single domain such as NYUv2~\cite{NYUv2} and KITTI~\cite{KITTIDC}, they often fail catastrophically on unseen sparse depth patterns or new datasets~\cite{ognidc,SpAgNet,VPP4DC}. As a result, users of downstream tasks such as view synthesis~\cite{dense_depth_prior_nerf} or 3D reconstruction~\cite{DC_app_3d_recon} have to train their own DC models on custom datasets, which is not only laborious but also could be infeasible if not enough RGB-D data are available for the test domain. This greatly limits the real-world applications of existing DC models.

This paper focuses on the most challenging and practical setting, \ie, \textit{zero-shot} generalization across different \textit{sparsity and sensors} with a \textit{single} model. This goal poses various challenges in the model design, due to the wide distribution of scene types and depth patterns that need to be covered. Therefore, we propose several novel designs in our \methodname\ framework to maximize its generalizability. The main contributions of this paper are as follows:
\begin{itemize}
    \item[--] We propose a novel layer named Multi-resolution Depth Integrator, and a novel Laplacian loss. The combination of them allows our method to deal with sparse depth maps of varying densities, even the extremely sparse ones.
    \item[--] We propose a scale normalization method together with a pipeline for generating synthetic sparse depth patterns, which allows us to do large-scale stable training on a mixture of datasets to enhance generalization.
    \item[--] Our method achieves state-of-the-art accuracy on seven datasets, reducing the error by as much as 43\% compared to baselines, and can be directly applied to view synthesis.
\end{itemize}
We now provide more details on the main components.

\noindent\textbf{Multi-Resolution Depth Integrator.} Recent work OGNI-DC~\cite{ognidc} shows impressive generalization capability with its Differentiable Depth Integrator design. However, a key limitation of OGNI-DC is its poor performance on extremely sparse input depth. We first study the cause of this limitation through theoretical analysis, and find that this is due to the error accumulation in the depth integration process. Based on that, we propose a Multi-resolution Depth Integrator, which allows explicit modeling of long-range depth relationships and significantly improves the performance.

\noindent\textbf{Laplacian Loss.} Existing DC models are trained with an $L_1$ or $L_2$ loss~\cite{completionformer,ognidc}. We find that these losses are dominated by the ambiguous regions (\eg, areas lacking sparse depth points), resulting in poor convergence. To resolve this issue, we propose a probability-based Laplacian loss, which allows our method to self-adaptively model the uncertainty during the training process, leading to better results.

\noindent\textbf{Scale Normalization.} DC models are usually trained on a single dataset (\eg, NYU or KITTI), making them over-fitted to a specific domain and resulting in poor generalization. While training monocular depth models on a mixture of datasets is a promising solution~\cite{midas,depth_anything_v2}, naive mixing leads to poor performance, due to the vastly different depth ranges. To this end, we introduce a novel scale normalization technique tailored to the DC task, which predicts in the $\log$-depth space and matches the median values across samples, resulting in good convergence on all types of scenes. 

\noindent\textbf{Large-Scale Training with Synthetic Depth Patterns.}
We train \methodname\ on 5 large-scale, high-quality datasets, covering indoor, outdoor, and urban scenes. In order to enhance the diversity of training data and to align the model with the real-world sensor distributions, we design a diverse set of synthetic depth patterns to generate the sparse depth maps, including LiDAR, SfM, and two types of noises. 

\noindent\textbf{State-of-the-Art Zero-Shot DC Accuracy.} We conduct extensive experiments across seven real-world datasets, including the conventional DC benchmarks KITTI~\cite{KITTIDC} and NYUv2~\cite{NYUv2}. We also test on two datasets with real SfM or VIO depth patterns (\ie, ETH3D~\cite{eth3d} and VOID~\cite{VOID}). Finally, since datasets with real sparse depth measurements are limited, we test on 4 standard monocular depth benchmarks with synthetically generated sparse depth patterns.

\methodname\ outperforms all baselines by a large margin. On the outdoor split of ETH3D~\cite{eth3d} with real SfM points, our model achieves RMSE=1.069, a 43\% reduction from the second best method Marigold~\cite{marigold}. On KITTI with 8-lines LiDAR, our model achieves a zero-shot MAE=0.597, even better than all methods trained on KITTI.

Finally, we show a practical application of \methodname\ on view synthesis with sparse input views. We train 3DGS~\cite{3dgs} with an auxiliary depth loss~\cite{dn-splatter}. The rendering quality improves greatly compared to the vanilla 3DGS (PSNR=20.38 vs 15.64) or using other depth supervisions~\cite{ZoeDepth,g2-monodepth}.

\begin{figure*}[ht]
    \centering
    \vspace{-8mm}
    \includegraphics[width=\textwidth]{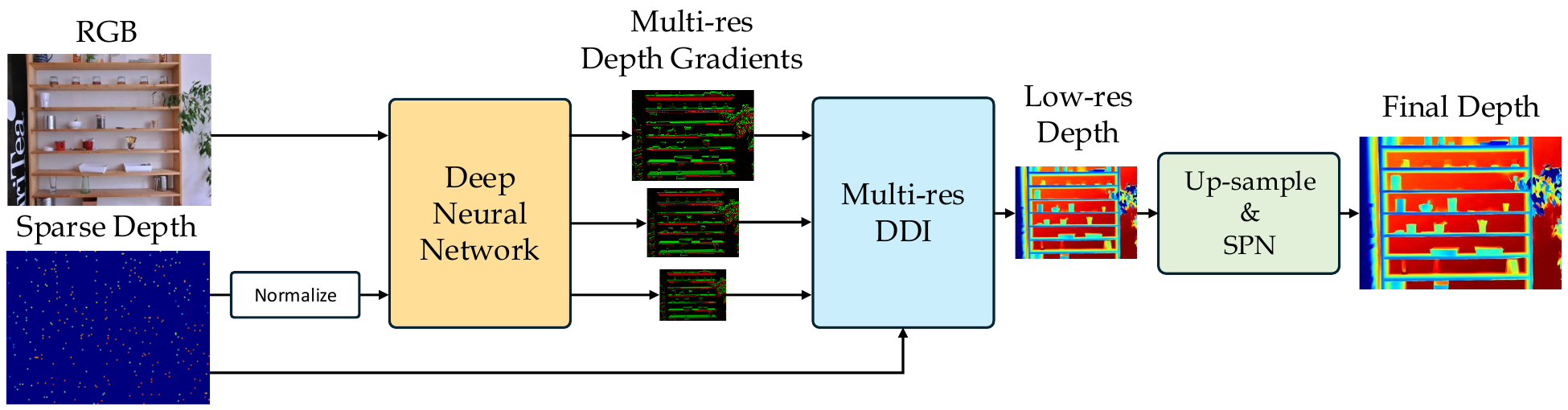}
    \vspace{-6mm}
    \caption{The overall pipeline of \methodname. The RGB image and the normalized sparse depth (\cref{sec:scale_normalization}) are fed into a neural network to produce a set of multi-resolution depth gradient maps. These depth gradient maps are integrated into a dense depth map with the Multi-resolution DDI (\cref{sec:multires_DDI}). Finally, the dense map is up-sampled and processed by an SPN~\cite{DySPN} to produce the final prediction (\cref{sec:implementation_details}).}
    \vspace{-6mm}
    \label{fig:pipeline}
\end{figure*}

\section{Related Work}
\noindent \textbf{Depth Completion Models.} In recent years, various deep-learning-based models have been proposed to tackle the DC task. Liu \etal~\cite{SPN} introduces the spatial propagation network (SPN), which iteratively propagates the initial predictions to its neighboring pixels through a set of learned combination weights. Different SPN variants have been proposed~\cite{CSPN,DSPN,CSPN++,GCSPN,NLSPN,DySPN}. LRRU~\cite{LRRU} and DFU~\cite{DFU} first use heuristics-based algorithms to pre-fill the depth maps and then learn to refine the initial predictions. BP-Net~\cite{BP-Net} employs a learned pre-processing stage to directly propagate the sparse depth points. Other works focus on improving the neural network architecture for DC models. CompletionFormer~\cite{completionformer} proposes a hybrid Transformer-CNN backbone suitable for the DC task. OGNI-DC~\cite{ognidc} first predicts depth gradients with a ConvGRU and uses a depth integration layer (DDI) to convert it into depth, and then performs iterative updates. Our model is based on OGNI-DC, but we propose novel designs including Multi-res DDI and the Laplacian loss, which significantly improve the performance of OGNI-DC on extremely sparse depth inputs. Moreover, all these methods focus on overfitting to a single type of sensor on a single dataset, in contrast to the dataset and sensor agnostic generalizable setting of our method.

\noindent\textbf{Generalization Across Depth Patterns.} Several works focus on sparsity-agnostic generalization. SpAgNet~\cite{SpAgNet} proposes a depth scaling and substitution module and can deal with very sparse depth inputs. Xu \etal~\cite{domain_agnostic_dc} utilizes a guidance map from a monocular depth network. VPP4DC~\cite{VPP4DC} finetunes a stereo matching network on a synthetic dataset by projecting the sparse depth points as virtual mosaic patterns onto the images. Depth Prompting~\cite{depth_prompting} starts from a monocular depth network and learns affinities from the sparse depth map for value propagation. While these methods can generalize across different sparsity, they are trained on a single dataset and cannot generalize beyond the training domain. In contrast, our \methodname\ is trained on a mixture of data and can generalize zero-shot to new datasets.

\noindent\textbf{Generalization Across Datasets.} Previous works use different settings for cross-dataset generalization. UniDC~\cite{UniDC} focuses on the few-shot setting (\ie, 1-100 labeled images), and utilizes depth foundation models with hyperbolic space for fast adaption. TTADC~\cite{test_time_adaptation} utilizes test-time adaptation (which requires a lot of unlabeled images) to close the domain gaps. Our \textit{zero-shot} setting is strictly more challenging, and also more practical because no data need to be collected from the target domain for fine-tuning.

More recently, G2-MonoDepth (G2-MD)~\cite{g2-monodepth} also focuses on the zero-shot setting and proposes to jointly solve monocular depth estimation and depth completion with one model by using a unified loss. However, G2-MD only tests with random sparse depth samples, which doesn't match the distribution of sensors in the real world. We do a large-scale comparison with G2-MD and show that our method consistently performs better than G2-MD on all datasets.

\smallskip \noindent \textbf{Generalizable Depth Estimation}. Various models~\cite{ZoeDepth,depth_pro,DepthAnything,depth_anything_v2,marigold,geowizard} have been proposed for generalizable depth estimation since MiDaS~\cite{midas}. Depth Anything v2~\cite{DepthAnything,depth_anything_v2} trains on pseudo-labels generated on unlabeled real images. Depth Pro~\cite{depth_pro} proposes a two-stage training strategy to first train on real images and then only on synthetic images for finer details. Marigold~\cite{marigold} and GeoWizard~\cite{geowizard} use pre-trained diffusion models as a strong prior.  Compared to them, our \methodname\ simplifies the training pipeline by training purely on synthetic images from scratch, but generalizes surprisingly well to real-world benchmarks. Furthermore, we show that the existing monocular depth models cannot effectively utilize the sparse depth information, and work poorly on the DC benchmarks.

\begin{figure*}[ht]
    \centering
    \vspace{-7mm}
    \includegraphics[width=\textwidth]{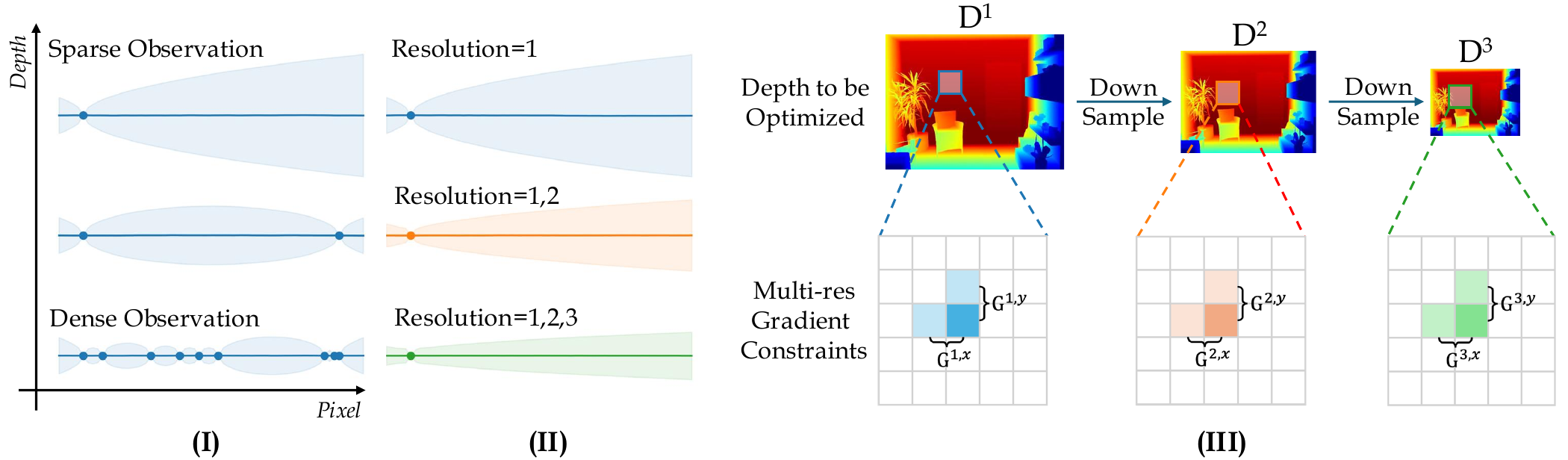}
    \vspace{-7mm}
    \caption{The Multi-resolution DDI reduces the error accumulation in the depth integration. In \textbf{(I)} and \textbf{(II)} we mark the pixels with known depth in dots, and show the mean and the 95\% confidence interval of the integrated depth, assuming the ground-truth depth gradients are $0$, and an i.i.d. Gaussian noise on their predictions. \textbf{(I)}: Noise accumulates in the integration process, especially obvious when the known depths are sparse. \textbf{(II)}: Our Multi-resolution DDI reduces the error, as it explicitly models the long-range depth dependencies. \textbf{(III)}: Multi-res DDI is implemented by down-sampling the optimization target depth map and computing the finite difference at each resolution.}
    \label{fig:multiresDDI-motivation}
    \vspace{-4mm}
\end{figure*}

\section{Methods} 

We define the task of depth completion as follows: the model takes an RGB image $\mathbf{I} \in \mathbb{R}^{3 \times H \times W}$, a sparse depth observation map $\mathbf{O} \in \mathbb{R}_+^{H \times W}$, and a valid mask $\mathbf{M} \in \{0,1\}^{H \times W}$ as input. $\mathbf{M}$ specifies the pixels with valid depth values in $\mathbf{O}$.
The model outputs a dense depth map with the same resolution as input: $\mathbf{\hat{D}} \in \mathbb{R}_+^{H \times W}$. In the rest of this paper, we use hat (\ $\hat{\cdot}$\ ) to denote the predicted values.

The overall pipeline of our method is shown in \cref{fig:pipeline}. In the
rest of this section, we introduce OGNI-DC and its limitations (\cref{sec:ogni-dc,sec:DDI_limitation}), the Multi-res DDI design
(\cref{sec:multires_DDI}), the Laplacian loss (\cref{sec:laplacian_loss}), the scale normalization (\cref{sec:scale_normalization}), and the training paradigm (\cref{sec:training_data}).

\subsection{Preliminaries: OGNI-DC}
\label{sec:ogni-dc}

OGNI-DC~\cite{ognidc} begins by using a deep neural network $F$ to predict a depth gradient map $\mathbf{\hat{G}}$. $\mathbf{\hat{G}}$ models the depth relationships between neighboring pixels (see \cref{fig:multiresDDI-motivation}.III):

\vspace{-2mm}
\begin{equation}
    \mathbf{\hat{G}} = [\mathbf{\hat{G}}^x, \mathbf{\hat{G}}^y] = F(\mathbf{I}, \mathbf{O}; \theta),
\end{equation}
where $\theta$ is the parameters of the neural network; $x$ and $y$ are the horizontal and the vertical direction, respectively.

The key component of OGNI-DC is a parameter-free custom layer named Differentiable Depth Integrator (DDI). DDI takes the depth gradient map and the sparse depth map as input, and outputs a dense depth map. This is achieved by solving a linear least squares problem involving constraints from both the sparse depths and the depth gradients:

\vspace{-4mm}
\begin{equation}
    \label{eq:optim_target}
    \mathbf{\hat{D}} = \argmin_{\mathbf{D}} \left(\alpha \cdot \mathcal{E}_O (\mathbf{D}, \mathbf{O},\mathbf{M} ) + \mathcal{E}_G (\mathbf{D}, \mathbf{\hat{G}})  \right),
\end{equation}
where $\alpha$ is a hyperparameter, and 
\begin{equation}
\label{eqn:ognidc-energy}
\begin{split}
    \mathcal{E}_O &:= \sum_{i,j}^{W,H} \mathbf{M}_{i,j} \cdot (\mathbf{D}_{i,j} - \mathbf{O}_{i,j})^2, \\
      \mathcal{E}_G &:=  \sum_{i,j}^{W,H} \left( \mathbf{G}^x_{i,j} - \mathbf{\hat{G}}^x_{i,j} \right) ^ 2
    + \left(\mathbf{G}^y_{i,j} - \mathbf{\hat{G}}^y_{i,j} \right) ^ 2,
\end{split}
\end{equation}
with $i,j$ being the pixel index; $\mathbf{G}^x$ and $\mathbf{G}^y$ being the finite differences: $\mathbf{G}^x_{i,j} := \mathbf{D}_{i,j} - \mathbf{D}_{i-1, j}$; $\mathbf{G}^y_{i,j} := \mathbf{D}_{i,j} - \mathbf{D}_{i, j-1}$.

Intuitively, $\mathcal{E}_O$ encourages the predicted depth to be consistent with the observed depth at valid locations, and $\mathcal{E}_G$ fills the missing areas with the learned depth gradients. DDI can be loosely understood as an integration process from known pixels to unknown ones. DDI alleviates the need for the neural network to learn an identity mapping for known pixels, thereby providing a strong inductive bias.

\subsection{Limitation of DDI on Extremely Sparse Depth}
\label{sec:DDI_limitation}

\begin{figure*}[ht]
    \centering
     \vspace{-9mm}
    \includegraphics[width=\textwidth]
    {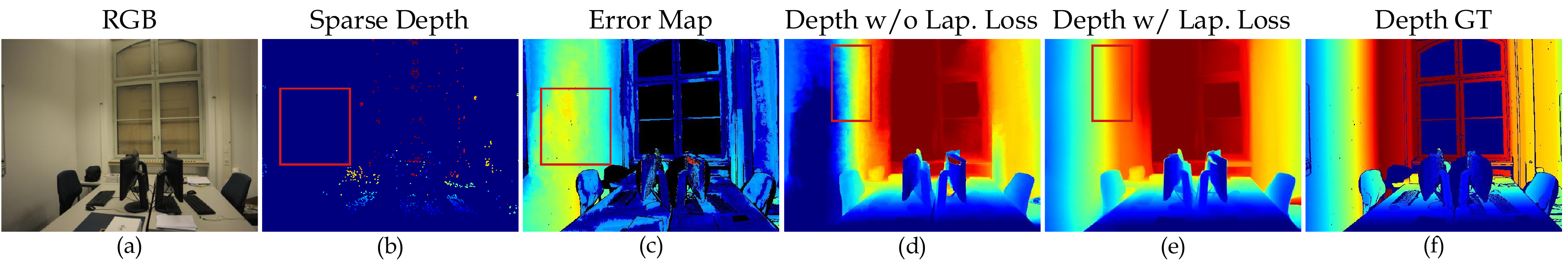}
    \vspace{-7mm}
    \caption{The sparse depth map from COLMAP~\cite{COLMAP_SfM} often has large holes on the textureless surfaces (b). These areas with high ambiguity dominate the $L_1$ training error (c). As a result, the model trained without the Laplacian loss does not converge well, producing artifacts in the depth map (d). In contrast, our model with the Laplacian loss generates a high-quality, smooth depth map (e).}
    \label{fig:probablity-loss}
    \vspace{-5.5mm}
\end{figure*}

While OGNI-DC achieves good generalization, it performs poorly when the depth observations are extremely sparse, \eg, only 5 points on NYUv2~\cite{ognidc}. This limitation causes problems in real-world applications: when the sparse depths are obtained from SfM, the texture-less surfaces often have no reliable correspondence (\cref{fig:probablity-loss}). Similarly, active sensors often fail to generate depth on transparent or metallic surfaces, leaving large blank areas in the depth maps~\cite{active_sensor_limitations,HAMMEER}. 

We examine the cause of this limitation, and find that it is due to the error accumulation in the long-range integration. To illustrate this, we simplify the problem into 1D and assume an i.i.d. Gaussian additive noise with variance $\sigma^2$ on the network's depth gradient prediction at pixel $i$:

\begin{equation}
    \mathbf{\hat{G}}_{i} = \mathbf{G}^{\text{gt}}_{i} + \mathbf{n}_i, \mathbf{n}_i \sim \mathcal{N}(0, \sigma^2).
\end{equation}

Assuming we only know the depth at pixel location $0$, $\mathbf{D}_0$, the predicted depth at location $n$ is obtained by integrating the gradient values from $0$ to $n$:

\vspace{-4mm}

\begin{equation}
    \mathbf{\hat{D}}_{n} = \mathbf{D}_0 + \sum_{i=1}^n \mathbf{\hat{G}}_{i} \sim \mathcal{N}(\mathbf{D}_0 + \sum_{i=1}^n \mathbf{G}^{\text{gt}}_{i}, n \cdot \sigma^2).
\end{equation}

The variance of $\mathbf{\hat{D}}_{n}$, $n \cdot \sigma^2$, increases linearly w.r.t. the distance to the nearest known pixel. It implies that the neural network's prediction error accumulates in the integration process, and the depth predictions are sensitive to the error in the depth gradient predictions when modeling long-range relationships. As illustrated in \cref{fig:multiresDDI-motivation}.I, when the observed depth map is relatively dense, the error accumulation is negligible. However, when the observed depths become sparser, the regions far from the observations become under-constrained and have a high depth prediction error. This explains why OGNI-DC performs badly on extremely sparse inputs.

\subsection{Multi-Resolution Depth Integration}
\label{sec:multires_DDI}

We propose a simple yet effective solution to enable DDI to overcome this limitation. We formulate a multi-resolution integration process, which jointly considers the depth relationships across multiple scales, reducing the integration error over long distances. Formally, we adjust the network to predict a set of depth gradient maps at different scales, where the resolution of each scale is different by a factor of 2, and $R$ is the total number of resolutions:

\vspace{-4mm}

\begin{equation}
    \{\mathbf{\hat{G}}^r\}_{r=1,\dots,R} = F(\mathbf{I}, \mathbf{O}; \theta), \mathbf{\hat{G}}^r \in \mathbb{R}^{2 \times H/2^{r-1}\times W/2^{r-1}}.
\end{equation}

We then extend the original DDI to incorporate multi-resolution depth gradients. Denote $\mathbf{D}$ to be the depth map to be optimized in \cref{eq:optim_target}. We down-sample $\mathbf{D}$ with a set of average-pooling layers:

\vspace{-4mm}
\begin{equation}
    \mathbf{D}^r = \operatorname{AvgPool2D}(\mathbf{D}, 2^{r-1}), r=1,\dots,R.
\end{equation}

We modify $\mathcal{E}_G$ in \cref{eqn:ognidc-energy} and define the multi-resolution depth gradients energy term as 

\vspace{-3mm}

\begin{equation}
      \mathcal{E}_G^R :=\sum_{r=1}^R \sum_{i,j}^{W,H} \left( \mathbf{G}^{r,x}_{i,j} - \mathbf{\hat{G}}^{r,x}_{i,j} \right) ^ 2
    + \left( \mathbf{G}^{r,y}_{i,j} - \mathbf{\hat{G}}^{r,y}_{i,j} \right) ^ 2,
\end{equation}
where $\mathbf{G}^{r, x}_{i,j} := \mathbf{D}^{r}_{i,j} - \mathbf{D}^{r}_{i-1, j}$; $\mathbf{G}^{r, y}_{i,j} := \mathbf{D}^{r}_{i,j} - \mathbf{D}^{r}_{i, j-1}$. Finally, we solve the linear least squares following \cref{eq:optim_target} to get the layer output $\mathbf{\hat{D}}$.

The computation of the multi-resolution constraints is illustrated in \cref{fig:multiresDDI-motivation}.III, and the benefit is demonstrated in  \cref{fig:multiresDDI-motivation}.II:
the error bound of the integrated depth is reduced greatly in the extremely sparse input case when using 3 resolutions, compared to the vanilla DDI with 1 resolution. Intuitively, Multi-resolution DDI achieves a better modeling of the global structure, as the steps required for integration are reduced from $n$ to $n/2^{R-1}$ for a pixel $n$ distance away from the nearest known pixel, and the local details are still preserved by the constraints at the finer resolutions.

\begin{figure}[t]
    \centering
    \vspace{-2mm}
    \includegraphics[width=\linewidth]{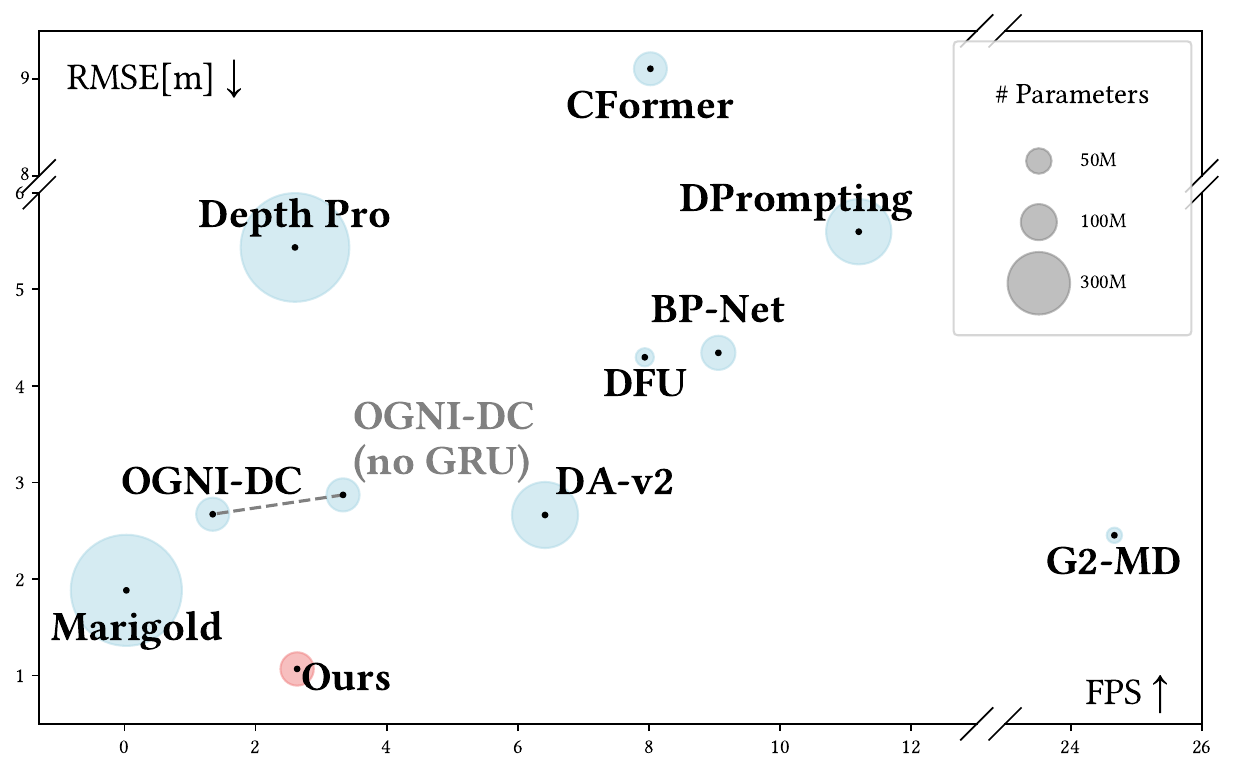}
    \vspace{-6.7mm}
    \caption{All numbers are benchmarked on the ETH3D~\cite{eth3d} outdoor split with SfM points. We use a single 3090 GPU and an image resolution of 480$\times640$. Our model achieves the best accuracy with a small model size (85M vs 907M for Depth Pro) and a competitive running speed (93$\times$ faster than Marigold~\cite{marigold} and 2$\times$ faster than OGNI-DC~\cite{ognidc}). As an ablation, we report the speed of OGNI-DC without its iterative GRU updates. Our multi-resolution design brings slightly higher latency (300ms vs 379ms, +26.6\%) compared to ``OGNI-DC (no GRU)", but much better accuracy.}
    \label{fig:fps_acc_curve}
        \vspace{-5mm}
\end{figure}

Note that the number of constraints decreases exponentially as the resolution increases. Therefore, the additional computation overhead is marginal compared to the vanilla DDI. A comparison of the inference speed and the parameter count of  \methodname\ against baselines is shown in \cref{fig:fps_acc_curve}. Our method is $2\times$ faster than OGNI-DC in inference.

\subsection{Laplacian Loss}
\label{sec:laplacian_loss}

As shown in \cref{fig:probablity-loss} (b), the sparse depth maps often contain large blank areas (``holes") with missing depth observation. Models typically make much larger errors in these areas due to the high ambiguity of depth values (\cref{fig:probablity-loss} (c)).

Therefore, when trained with an $L_1$ loss, the model focuses on optimizing the high-ambiguity regions to capture the global structure accurately. The local details are thus not well optimized, leaving artifacts in the predicted depth maps, as shown in \cref{fig:probablity-loss} (d).

We propose to use a probability-based loss to explicitly model the uncertainty of the depth prediction to achieve a smoother result. Specifically, rather than predicting a single depth value, the model predicts the mean $\mathbf{\hat{D}}$ and a per-pixel scale parameter $b$ of the Laplacian distribution, and we use its negative log-likelihood as the Laplacian loss:
\vspace{-0.5mm}
\begin{equation}
    L_{Lap}(\mathbf{D}^{\text{gt}}, \mathbf{\hat{D}}, b) = \log(2b) + |\mathbf{D}^{\text{gt}} - \mathbf{\hat{D}}|/{b}.
\end{equation}

\vspace{-0.5mm}

$L_{Lap}$ can be viewed as a probabilistic variant of $L_1$. We don't use an $L_2$ or Gaussian loss because we find $L_2$ more sensitive to outliers and make training less stable.

Although training with $L_{Lap}$ alone produces smoother results, it reduces the model’s ability to handle noise in the sparse depth map, as it can cheat by predicting large uncertainties (See Appendix \cref{sec:appendix_laplacian_loss}). We find that combining $L_{Lap}$ with the $L_1$ loss yields the best results. We also adopt the gradient-matching loss $L_{gm}$ proposed in the monocular depth estimation~\cite{megadepth,midas,DepthAnything} on the predicted depth map.

The final loss can be written as:

\vspace{-2mm}
\begin{equation}
    L = L_1 + 0.5 \cdot  L_{Lap} + 2.0 \cdot L_{gm}.
\end{equation}

While probability-based losses have been used in previous works on various tasks~\cite{weight_uncertainty,human_pose_uncertainty,LOFTER,PDC-Net,SEA_RAFT}, we are the first to apply it to the task of depth estimation or depth completion, and prove its usefulness through experiments.

\subsection{Scale Normalization}
\label{sec:scale_normalization}
We desire our DC model to work well across a large variety of scenes, which have a large variation in the depth scale, \eg, $<1m$ for indoors and $>100m$ for urban scenes. 

Several problems will occur if we naively process depth in the metric space like previous DC methods~\cite{ognidc,completionformer}. 1) \textit{Model Capacity}: since the sparse depth map is part of the neural network input, the network has to learn to process a wide value range, posing challenges to the network capacity. 2) \textit{Unbalance Among Datasets}: the commonly used $L_1$ or $L_2$ losses incur a larger penalty on larger depth values when the relative errors are the same. Therefore, the training loss focuses more on outdoor scenes, which is undesirable. 3) \textit{Scale Ambiguity in SfM}: SfM algorithms can only reconstruct scenes up to an arbitrary global scale~\cite{COLMAP_SfM}. Recovering metric depth is often impossible and sometimes unnecessary for applications such as view synthesis~\cite{dense_depth_prior_nerf}. 

In monocular depth estimation, the scale ambiguity is often resolved by using the scale-invariant
loss~\cite{midas,DepthAnything,depth_anything_v2}. However, this won't work in DC, because instead of allowing the
model to choose an arbitrary scale to predict, we require the
scale of the output depth to be \textit{equivariant} to the scale of the input
sparse depth in DC. For an arbitrary scale factor $\beta$, scale equivariance is formally defined as:

\vspace{-1mm}
\begin{equation}
    \mathbf{\hat{D}}(\mathbf{I}, \beta \cdot \mathbf{O}) = \beta \cdot \mathbf{\hat{D}}(\mathbf{I}, \mathbf{O}), \forall \beta \in \mathbb{R}_{+}.
\end{equation}

To address the scale issue, we propose a scale normalization technique tailored to DC. First, we convert all depth into the $\log$ space, where the arbitrary multiplicative scale factor becomes additive, making it suitable for the linear formulation of DDI. Second, we normalize the input sparse depth map by its median value, so that the value ranges are matched between the indoor and outdoor scenes:

\vspace{-4mm}

\begin{equation}
    \mathbf{\hat{G}} = F(\mathbf{I}, \mathbf{\Tilde{O}}; \theta), \mathbf{\Tilde{O}} = \log(\mathbf{O}) - \log(\operatorname{median}(\mathbf{O})).
    \vspace{-1mm}
\end{equation}

As illustrated in \cref{fig:pipeline}, only the network input is normalized, but \textit{not} the sparse depth used in DDI. Therefore, the original scale of $\mathbf{O}$ is preserved in the final output through DDI, and our model achieves \textit{guaranteed} scale equivariance. See Appendix \cref{sec:appendix_scale_equivariance_proof} for proofs.

\subsection{Large-Scale Training with Synthetic Patterns}
\label{sec:training_data}

We train our model on a collection of 5 synthetic datasets, covering indoor, outdoor, and urban scenes, with a total of 573K images. The details of the datasets used for training are shown in \cref{tab:datasets}. We choose synthetic datasets because real-world datasets with high-quality depth ground-truth are very limited, and we find that mixing real-world datasets (\eg, NYU) for training produces blurry results, as shown in Appendix \cref{sec:appendix_data_ablations}. While we don't use the complicated two-stage strategy mixing unlabeled real-world data~\cite{depth_anything_v2,depth_pro}, we find that synthetic data alone yields surprisingly good results on real-world benchmarks. 

During training, the sparse depth maps are synthetically generated by sub-sampling the dense ground-truth. Previous works such as G2-MD~\cite{g2-monodepth} use random samples, which align poorly with the sparse point distributions of the real sensors. We instead design two kinds of synthetic sparse patterns: 1) \textit{SfM}: sparse points are sampled at the SIFT~\cite{SIFT} keypoints. 2) \textit{LiDAR}: we simulate a random 4-128 lines LiDAR with angle and shift variations. 

Additionally, we simulate two types of noise for generating the sparse depth map. 1) \textit{Outliers}: this is common in the COLMAP output due to mismatched keypoints. We simulate the outliers by randomly sampling depth values within the scene depth bounds. 2) \textit{Boundary Noise}: blended foreground and background depth points near object boundaries occur due to viewpoint differences between the LiDAR and RGB camera have been observed in \cite{KITTI_noisy_1,deeplidar}. We simulate it by projecting the depth map to a virtual neighboring view, inpainting the holes, sampling sparse depth, and projecting back. See Appendix \cref{sec:appendix_implementation_details} for more details.

\section{Experiments}
\label{sec:experiments}

\subsection{Evaluation Datasets}

\label{sec:robust_dc_benchmark}

We test on 7 real-world datasets to show the effectiveness of our method, as shown in~\cref{tab:datasets}. For NYUv2, we use the standard test set and cropping following~\cite{completionformer,ognidc} and test on two different densities (500/50 points). For KITTI, we use its official validation set and follow~\cite{twin_surface_dc} to sub-sample the original 64-lines lidar into sparser inputs. We use the original test set of VOID~\cite{VOID} with 3 densities (1500/500/150).

To test the model's capability on dealing with SfM inputs, we use the ETH3D~\cite{eth3d} dataset and project the COLMAP SfM sparse depth points into the image space to get sparse depth maps. We call this dataset ETH3D-SfM.

Finally, since the datasets with sparse depth from real sensors are limited, we utilize 4 benchmarks commonly used in monocular depth estimation (iBims, ARKitScenes, ETH3D, DIODE) and generate sparse depth patterns synthetically. We sample random points with different densities [0.7\%/0.1\%/0.03\%]; add [5\%/10\%] noise on top of the 0.7\% density; sample depth at the [SIFT~\cite{SIFT}/ORB~\cite{ORB}] keypoints; construct synthetic LiDAR points with [64/16/8] scanning lines. This synthetic subset helps us understand how these factors alone affect the model performance.

\subsection{Implementation Details}
\label{sec:implementation_details}

We use CompletionFormer~\cite{completionformer} as the backbone, and 3 resolutions for the DDI. The DDI generates an intermediate depth map at the $1/4$ resolution, which is refined by a convex up-sampling layer~\cite{RAFT} and a DySPN~\cite{DySPN}, following OGNI-DC~\cite{ognidc}. Compared to \cite{ognidc}, we remove the iterative updates with ConvGRU because we find it not helpful for performance when trained on large-scale datasets.

We train \methodname\ on 10$\times$48GB GPUs, with an effective batch size of 60. In each epoch, we randomly sample 25K images from each dataset. The model is trained for 72 epochs, which takes about 6 days in total. Additional details are provided in Appendix \cref{sec:appendix_implementation_details}.

 \begin{table}[t]
\setlength{\tabcolsep}{1.2mm}
\small
\centering
        \caption{Detailed statistics of the datasets used in this paper. \textsuperscript{\textdagger}ARKitScenes~\cite{arkitscenes} has 450K images in total. We randomly sample 800 images from its validation split for faster evaluation.}
        \vspace{-3mm}
        \label{tab:datasets}
            \begin{tabular}{@{}clccc@{}}
              \hline
              \noalign{\vskip 0.2mm}  
   Split & Dataset Name & Size & Sparse Depth & Scene Type\\
   
  \hline \hline  
  \multirow{5}{*}{Training} & Hypersim~\cite{Hypersim} & 66K & \multirow{5}{*}{{\begin{tabular}{@{}c@{}}{Synthetic} \\ {(\cref{sec:training_data})} \end{tabular}}} & Indoor \\
  & IRS~\cite{IRS} & 60K & & Indoor \\
  & Tartanair~\cite{TartanAir} & 307K & & In/Out \\
  & BlendedMVS~\cite{BlendedMVS} & 115K & & Misc \\
  & Virtual KITTI~\cite{vkitti} & 25K & & Urban \\
    \hline 
  \multirow{7}{*}{Testing} & iBims~\cite{ibims} & 100 & Synthetic & Indoor \\
  & ARKitScenes~\cite{arkitscenes} & 800\textsuperscript{\textdagger} & Synthetic & Indoor \\
  & VOID~\cite{VOID} & 2400 & SfM & Indoor \\
  & NYUv2~\cite{NYUv2} & 652 & Synthetic & Indoor \\
  & DIODE~\cite{diode} & 771 & Synthetic & In/Out \\
  & ETH3D~\cite{eth3d} & 454 & SfM & In/Out \\
  & KITTIDC~\cite{KITTIDC} & 1000 & LiDAR & Urban \\
  
  \hline
  \end{tabular}
  \vspace{-5mm}
\end{table}

\begin{table*}[t]
\setlength{\tabcolsep}{1.16mm}
\small
\centering
        \vspace{-9mm}
        \caption{Large-scale testing on diverse datasets with synthetic sparse depth patterns. The \colorbox{red!50}{1\textsuperscript{st}}, \colorbox{orange!50}{2\textsuperscript{nd}}, \colorbox{yellow!50}{3\textsuperscript{rd}} place methods are marked accordingly. Results are averaged on 4 datasets (\cref{sec:robust_dc_benchmark}). Definitions of the sparse depth patterns can be found in \cref{sec:robust_dc_benchmark}.}
        \vspace{-3mm}
        \label{tab:results_virtual_pattern}
            \begin{tabular}{@{}lP{1.2cm}P{1.2cm}P{1.2cm}P{1.2cm}P{1.2cm}P{1.2cm}P{1.2cm}P{1.2cm}P{1.2cm}P{1.2cm}@{}}
              \hline
               \noalign{\vskip 0.2mm}  
  \multirow{2}{*}{Methods} &
  \multicolumn{2}{c}{0.7\%} &
  \multicolumn{2}{c}{0.1\%} &
  \multicolumn{2}{c}{0.03\%} &
  \multicolumn{2}{c}{5\% Noise} & 
  \multicolumn{2}{c}{10\% Noise}
  \\
  \cmidrule(lr){2-3} \cmidrule(lr){4-5} \cmidrule(lr){6-7} \cmidrule(lr){8-9} \cmidrule(lr){10-11}   
   &
  RMSE & REL &
   RMSE & REL &
   RMSE & REL & 
  RMSE & REL &
   RMSE &  REL \\
  \hline \hline  
  Depth Pro~\cite{depth_pro} & 0.938 &	0.259	& 0.938	& 0.259 &	0.938	& 0.259	& 0.938	& 0.259	& 0.938	& 0.259 \\
  DepthAnythingv2~\cite{depth_anything_v2} & 0.818	&	0.066	&	0.834	&	\cellcolor{yellow!50} 0.067	&	0.845	&	\cellcolor{orange!50} 0.068	&	1.299	&	0.238	&	1.922	&	0.380 \\
  Marigold~\cite{marigold} &  0.367	&	0.081	&	0.373	&	0.082	&	\cellcolor{orange!50} 0.384	&	\cellcolor{yellow!50} 0.084	&	0.379	&	0.083	&	0.406	&	0.091 \\
  CompletionFormer~\cite{completionformer} & 0.434 & 0.225 & 1.227 & 0.586 & 1.755 & 0.826 & 0.506 & 0.236 & 0.565 & 0.249 \\
  DFU~\cite{DFU} & 1.629 & 0.798 & 2.986 & 1.481 & 4.447 & 2.277 & 1.697 & 0.805 & 1.759 & 0.813 \\
  BP-Net~\cite{BP-Net} & 0.361 & 0.044 & 0.898 & 0.185 & 1.147 & 0.257 & 0.418 & 0.058 & 0.478 & 0.076 \\
  OGNI-DC~\cite{ognidc} & \cellcolor{yellow!50} 0.187 & \cellcolor{yellow!50} 0.018 & \cellcolor{yellow!50} 0.355 & 0.068 & 0.557 & 0.143 & \cellcolor{yellow!50} 0.265 & \cellcolor{yellow!50} 0.029 & \cellcolor{yellow!50} 0.333 & \cellcolor{yellow!50} 0.041 \\
  G2-MonoDepth~\cite{g2-monodepth} &  \cellcolor{orange!50} 0.168 & \cellcolor{orange!50} 0.015 & \cellcolor{orange!50} 0.280 & \cellcolor{orange!50} 0.041 & \cellcolor{yellow!50} 0.434 & 0.094 & \cellcolor{orange!50} 0.193 & \cellcolor{orange!50} 0.016 & \cellcolor{orange!50} 0.214 & \cellcolor{orange!50} 0.018 \\
  
  \hline
  \textbf{Ours} & \cellcolor{red!50} 0.135	& \cellcolor{red!50} 0.010	&  \cellcolor{red!50} 0.211	&  \cellcolor{red!50}0.020	&  \cellcolor{red!50}0.289	&  \cellcolor{red!50}0.034 &	\cellcolor{red!50} 0.141 & \cellcolor{red!50} 0.010	& \cellcolor{red!50} 0.147	& \cellcolor{red!50} 0.011 \\

  \hline
  \noalign{\vskip 0.5mm}
  \hline
  \multirow{2}{*}{Methods} &
  \multicolumn{2}{c}{ORB~\cite{ORB}} &
  \multicolumn{2}{c}{SIFT~\cite{SIFT}} &
  \multicolumn{2}{c}{LiDAR-64-Lines} &
  \multicolumn{2}{c}{LiDAR-16-Lines} & 
  \multicolumn{2}{c}{LiDAR-8-Lines} \\
 \cmidrule(lr){2-3} \cmidrule(lr){4-5} \cmidrule(lr){6-7} \cmidrule(lr){8-9} \cmidrule(lr){10-11}   
  &
   RMSE & REL &
   RMSE & REL &
   RMSE & REL & 
   RMSE & REL &
   RMSE & REL \\
  \hline\hline
  Depth Pro~\cite{depth_pro} & 0.938 &	0.259	& 0.938	& 0.259 &	0.938	& 0.259	& 0.938	& 0.259	& 0.938	& 0.259 \\
  DepthAnythingv2~\cite{depth_anything_v2} & 1.343 &	0.569&		0.871&		0.299	&	0.974	&	0.110&		0.800&		0.065	& 0.822&	\cellcolor{yellow!50}	0.068 \\
  Marigold~\cite{marigold} & \cellcolor{yellow!50} 0.467	& \cellcolor{yellow!50}	0.140	&	\cellcolor{yellow!50} 0.453	&	\cellcolor{yellow!50} 0.136	&	0.367	&	0.081	&	0.369	&	0.082	&	\cellcolor{yellow!50} 0.378	&	0.082 \\
  CompletionFormer~\cite{completionformer} & 1.452 & 0.553 & 1.396 & 0.586 & 0.303 & 0.116 & 0.465 & 0.184 & 1.140 & 0.469  \\
  DFU~\cite{DFU} & 4.033 & 2.090 & 4.107 & 2.130 & 2.003 & 0.940 & 1.862 & 0.971 & 3.031 & 1.561 \\
  BP-Net~\cite{BP-Net} & 1.105 & 0.304 & 1.057 & 0.299 & 0.296 & 0.033 & 0.531 & 0.078 & 0.935 & 0.195 \\
  OGNI-DC~\cite{ognidc} & 0.639 & 0.179 & 0.524 & 0.151 & \cellcolor{yellow!50} 0.162 & \cellcolor{yellow!50} 0.014 & \cellcolor{yellow!50} 0.247 & \cellcolor{yellow!50} 0.033 & 0.415 & 0.085 \\
  G2-MonoDepth~\cite{g2-monodepth} & \cellcolor{orange!50} 0.427 & \cellcolor{orange!50} 0.110 & \cellcolor{orange!50} 0.391 & \cellcolor{orange!50} 0.104 & \cellcolor{orange!50} 0.143 & \cellcolor{orange!50} 0.012 & \cellcolor{orange!50} 0.217 & \cellcolor{orange!50} 0.024 & \cellcolor{orange!50} 0.306 & \cellcolor{orange!50} 0.042 \\

  \hline
  \textbf{Ours} & \cellcolor{red!50} 0.247 & \cellcolor{red!50} 0.045 & \cellcolor{red!50} 0.211 & \cellcolor{red!50} 0.037 & \cellcolor{red!50} 0.121 & \cellcolor{red!50} 0.008 & \cellcolor{red!50} 0.164 & \cellcolor{red!50} 0.014 & \cellcolor{red!50} 0.231 & \cellcolor{red!50} 0.023 \\
  \hline
  \end{tabular}
\end{table*}

\begin{table*}[t]
\setlength{\tabcolsep}{0.94mm}
        \small
\centering
        \vspace{-1.5mm}
        \caption{Results on ETH-3D, KITTI, and NYUv2. Numbers in \textcolor{gray}{gray} are trained on KITTI/NYU, and are thus excluded from the ranking.}
        \vspace{-3mm}
        \label{tab:results_real_pattern}
            \begin{tabular}{@{}lP{0.91cm}P{0.91cm}P{0.91cm}P{0.91cm}P{0.2cm}P{1.6cm}P{0.91cm}P{0.91cm}P{0.91cm}P{0.91cm}P{0.91cm}P{0.91cm}P{0.91cm}P{0.91cm}@{}}
              \hline 
  \multirow{2}{*}{Methods} &
  \multicolumn{2}{c}{ETH-SfM-Out} &
  \multicolumn{2}{c}{ETH-SfM-In} & 
  & &
  \multicolumn{2}{c}{KITTI-64Line} &
  \multicolumn{2}{c}{KITTI-8Line} &
  \multicolumn{2}{c}{NYU-500Pt} &
  \multicolumn{2}{c}{NYU-50Pt}
  \\
  \cmidrule(lr){2-3} \cmidrule(lr){4-5}
  \cmidrule(lr){8-9}
  \cmidrule(lr){10-11}
  \cmidrule(lr){12-13}
  \cmidrule(lr){14-15}
  & 
  RMSE & REL &
  RMSE & REL &
  & &
  RMSE & MAE & 
  RMSE & MAE &
  RMSE & REL &
  RMSE & REL
  \\
  \hline \hline  
    CFormer~\cite{completionformer} & 9.108 & 1.215 & 2.088 & 0.229 
    & \multicolumn{2}{c}{\multirow{5}{*}{\begin{tabular}{@{}c@{}}{Trained on} \\ {KITTI/NYU} \end{tabular}}}
    & \textcolor{gray}{0.741} & \textcolor{gray}{0.195} & \textcolor{gray}{3.650} & \textcolor{gray}{1.701} & \textcolor{gray}{0.090} & \textcolor{gray}{0.012} & \textcolor{gray}{0.707} & \textcolor{gray}{0.181} \\
  DFU~\cite{DFU} & 4.296 & 0.588 & 3.572 & 1.105 &
  & &
  \textcolor{gray}{0.713} & \textcolor{gray}{0.186} & \textcolor{gray}{3.269} & \textcolor{gray}{1.468} & \textcolor{gray}{0.091} & \textcolor{gray}{0.011} & \textcolor{gray}{-} & \textcolor{gray}{-} \\
  BP-Net~\cite{BP-Net} & 4.342 & 0.339 & 1.664 & 0.301 &
  &
  & \textcolor{gray}{0.784} & \textcolor{gray}{0.011} & \textcolor{gray}{2.391} & \textcolor{gray}{0.953} & \textcolor{gray}{0.089} & \textcolor{gray}{0.012} & \textcolor{gray}{0.609} & \textcolor{gray}{0.157} \\
  DPromting~\cite{depth_prompting} & 5.596 & 0.846 & 1.306 & 0.269 &
  & &
  \textcolor{gray}{1.078} & \textcolor{gray}{0.324} & \textcolor{gray}{1.791} & \textcolor{gray}{0.634} &
  \textcolor{gray}{0.105} & \textcolor{gray}{0.015} & \textcolor{gray}{0.213} & \textcolor{gray}{0.043}
  \\
  
  OGNI-DC~\cite{ognidc} & 2.671 & 0.268 & 1.108 & 0.181 & 
  & &
  \textcolor{gray}{0.750} & \textcolor{gray}{0.193} & \textcolor{gray}{2.363} & \textcolor{gray}{0.777} & 
  \textcolor{gray}{0.089} & \textcolor{gray}{0.012} &
  \textcolor{gray}{0.207} & \textcolor{gray}{0.038} \\
  \hhline{>{\arrayrulecolor [gray]{1.0}}------>{\arrayrulecolor {black}}---------}
  Depth Pro~\cite{depth_pro} & 5.433 & 0.441 & 0.928 & 0.208 & \multicolumn{2}{c}{\multirow{5}{*}{Zero-shot}} 
  & 4.893 & 3.233 & 4.893 & 3.233 & \cellcolor{yellow!50} 0.266 & 0.062 & \cellcolor{orange!50} 0.266 & \cellcolor{yellow!50} 0.062 \\
  DA-v2~\cite{depth_anything_v2} & 2.663 & \cellcolor{orange!50} 0.082 & \cellcolor{red!50} 0.592 &  \cellcolor{red!50} 0.065 & &
  & 4.561 & 1.925 & 4.689 & 1.951 & 0.309 & \cellcolor{yellow!50} 0.061 & 0.330 & 0.063 \\
  Marigold~\cite{marigold} & \cellcolor{orange!50} 1.883 & 0.252 & \cellcolor{yellow!50} 0.627 & \cellcolor{yellow!50} 0.152 & & 
  & \cellcolor{yellow!50} 3.462 & \cellcolor{yellow!50} 1.911 & \cellcolor{yellow!50} 3.498 & \cellcolor{yellow!50} 1.939 & 0.426 & 0.115 & 0.436 & 0.118 \\
  G2-MD~\cite{g2-monodepth} & \cellcolor{yellow!50} 2.453 & \cellcolor{yellow!50} 0.153 & 1.068 & 0.164 & &
  & \cellcolor{orange!50} 1.612 & \cellcolor{orange!50} 0.376 & \cellcolor{orange!50} 2.769 & \cellcolor{orange!50} 0.901 & \cellcolor{orange!50} 0.122 & \cellcolor{orange!50} 0.017 & \cellcolor{yellow!50} 0.286 & \cellcolor{orange!50} 0.056 \\

  \hhline{>{\arrayrulecolor {black}}----->{\arrayrulecolor [gray]{1.0}}-->{\arrayrulecolor {black}}--------}
  \textbf{Ours} & \cellcolor{red!50} 1.069 & \cellcolor{red!50} 0.053 & \cellcolor{orange!50} 0.605 & \cellcolor{orange!50} 0.090  & & 
  & \cellcolor{red!50} 1.191 & \cellcolor{red!50} 0.270 & \cellcolor{red!50} 2.058 & \cellcolor{red!50} 0.597 & \cellcolor{red!50} 0.111 & \cellcolor{red!50} 0.014 & \cellcolor{red!50} 0.225 & \cellcolor{red!50} 0.041  \\
  \hline
  \end{tabular}
  \vspace{-6mm}
\end{table*}

\begin{table}[ht]
\setlength{\tabcolsep}{1.74mm}
\small
\centering
\vspace{-2mm}
        \caption{Ablation studies on the validation set. Res=1 is the vanilla DDI in OGNI-DC. ``Synthetic" means the SfM+LiDAR patterns.}
        \vspace{-2mm}
        \label{tab:ablation}
            \begin{tabular}{@{}clccccc@{}}
              \hline 
  \noalign{\vskip 0.5mm}
   &
   \multirow{2}{*}{Methods} &   \multicolumn{2}{c}{ETH3D-SfM} &
  \multicolumn{2}{c}{KITTI-64} &  \\
  \cmidrule(lr{1em}){3-4} \cmidrule(lr{1em}){5-6}
  & &
  RMSE & REL &
  RMSE & MAE \\
  
  \hline \hline  
  \noalign{\vskip 0.5mm}
  \multirow{3}{*}{{\begin{tabular}{@{}c@{}}{Multi-res} \\ {DDI} \end{tabular}}} & DDI, Res=1 & 0.595 & 0.086 & 1.210 & 0.275 \\
  & DDI, Res=1,2 & 0.489 & 0.069 & 1.218 & 0.275 \\
  & \textBF{DDI, Res=1,2,3} & \textBF{0.459} & \textBF{0.064} & \textBF{1.188} & \textBF{0.267} \\
  \hline
  \noalign{\vskip 0.5mm}
  \multirow{4}{*}{Losses} & $L_1$ & 0.666 & 0.080 & 1.234 & 0.280 \\
  & $L_1$+$L_{Lap}$ & 0.598 & 0.083 & 1.224 & 0.278 \\
  & $L_1$+$L_{gm}$  & 0.547 & 0.082 & \textBF{1.155} & 0.282 \\
  & $\bm{L_1}$+$\bm{L_{Lap}}$+$\bm{L_{gm}}$ & \textBF{0.490} & \textBF{0.076} & 1.173 & \textBF{0.277} \\
  \hline
  \noalign{\vskip 0.5mm}
  \multirow{3}{*}{{\begin{tabular}{@{}c@{}}{Depth} \\ {Space} \end{tabular}}} & Linear & 0.886 & 0.155 & 1.289 & 0.305 \\
  & Log & 0.627 & 0.103 & 1.293 & 0.308 \\
  & \textBF{Log+Normalize} & \textBF{0.490} & \textBF{0.076} & \textBF{1.173} & \textBF{0.278} \\
  \hline
  \noalign{\vskip 0.5mm}
  \multirow{3}{*}{{\begin{tabular}{@{}c@{}}{Training} \\ {Pattern} \\ \end{tabular}}} & Random & 0.714 & 0.117 & 1.490 & 0.342 \\
  & Rand.+Synthetic & 0.647 & 0.089 & 1.402 & 0.336 \\
  & \textBF{Rand.+Syn.+Noise} & \textBF{0.490} & \textBF{0.076} & \textBF{1.173} & \textBF{0.278}
   \\
  \hline
  \end{tabular}
\end{table}

\begin{table}[ht]
\setlength{\tabcolsep}{1.18mm}
\small
\centering
        \vspace{-2mm}
        \caption{Results on the VOID~\cite{VOID} dataset under three densities.}
        \vspace{-3mm}
        \label{tab:results_void}
            \begin{tabular}{@{}lP{0.82cm}P{0.82cm}P{0.82cm}P{0.82cm}P{0.82cm}P{0.82cm}@{}}
              \hline
              \noalign{\vskip 0.5mm} 
   \multirow{2}{*}{Methods} &   \multicolumn{2}{c}{VOID-1500} &
  \multicolumn{2}{c}{VOID-500} & \multicolumn{2}{c}{VOID-150}  \\
  \cmidrule(lr{0.5em}){2-3} \cmidrule(lr{0.5em}){4-5} \cmidrule(lr{0.5em}){6-7}
  &
  RMSE & MAE &
  RMSE & MAE &
  RMSE & MAE \\
  \hline \hline  
    Depth Pro~\cite{depth_pro} & 0.734 & 0.385 & 0.697 & 0.373 & 0.758 & 0.392 \\
      DA-v2~\cite{depth_anything_v2} & 0.605 & 0.209 & \cellcolor{yellow!50} 0.582 & 0.209 & \cellcolor{red!50} 0.644 & \cellcolor{orange!50} 0.230 \\
  Marigold~\cite{marigold} & 0.630 & 0.240 & 0.607 & 0.241 & \cellcolor{yellow!50} 0.673 & 0.263 \\
  CFormer~\cite{completionformer} & 0.726 & 0.261 & 0.821 & 0.385 & 0.956 & 0.487 \\
  DFU~\cite{DFU} & 3.222 & 2.297 & 3.628 & 2.648 & 4.521 & 3.356 \\
  BP-Net~\cite{BP-Net} & 0.738 & 0.268 & 0.790 & 0.369 & 0.934 & 0.470 \\
  DPromting~\cite{depth_prompting} & 0.779 & 0.373 & 0.754 & 0.373 & 0.820 & 0.398 \\
  OGNI-DC~\cite{ognidc} & \cellcolor{yellow!50} 0.593 & \cellcolor{yellow!50} 0.175 & 0.589 & \cellcolor{yellow!50} 0.198 & 0.693 & 0.261 \\
  G2-MD~\cite{g2-monodepth} & \cellcolor{orange!50} 0.568 & \cellcolor{orange!50} 0.159 & \cellcolor{orange!50} 0.574 & \cellcolor{orange!50} 0.182 & 0.691 & \cellcolor{yellow!50} 0.247 \\

  \hline
  \textbf{Ours} & \cellcolor{red!50} 0.555 & \cellcolor{red!50} 0.150 & \cellcolor{red!50} 0.551 & \cellcolor{red!50} 0.164 & \cellcolor{orange!50} 0.650 & \cellcolor{red!50} 0.211 \\
  \hline
  \end{tabular}
  \vspace{-6mm}
\end{table}

\subsection{Baselines}

We compare against state-of-the-art DC baselines CompletionFormer~\cite{completionformer} (CFormer), DFU~\cite{DFU}, BP-Net~\cite{BP-Net}, OGNI-DC~\cite{ognidc}, Depth Promting~\cite{depth_prompting} (DPromting), and G2-MonoDepth~\cite{g2-monodepth} (G2-MD). We also compare against the generalizable metric depth (Depth Pro~\cite{depth_pro}) and affine-invariant depth models (DepthAnythingv2~\cite{depth_anything_v2} (DAv2) and Marigold~\cite{marigold}), for which we estimate the global scale and shift under the best alignment with the sparse depth. 

\subsection{Results on KITTIDC and NYUv2}

Results are shown in \cref{tab:results_real_pattern}. While tested zero-shot, \methodname\ even outperforms all DC methods trained on KITTI in terms of MAE (MAE=0.597 vs 0.634 for DPromting~\cite{ognidc}) on the 8-lines input. On the 64-lines input, ours works much better than all other methods tested zero-shot (RMSE=1.191 vs 1.612 for G2-MD~\cite{g2-monodepth}, a 26\% reduction).

On NYUv2 (\cref{tab:results_real_pattern}), \methodname\ works better than all other zero-shot methods on different densities, achieving comparable performance with models trained on NYU (500 points, RMSE=0.111 vs 0.089 for OGNI-DC~\cite{ognidc}).

\vspace{-0mm}
\subsection{Results on the VOID and ETH3D-SfM}

VOID~\cite{VOID} has ground-truth depth collected with an Intel RealSense camera, and the sparse depths come from a visual-inertial odometry system with three different sparsity levels. As shown in \cref{tab:results_void}, our method outperforms all baselines by a large margin for denser inputs (1500/500), and is very close to DAv2~\cite{depth_anything_v2} on the 150 points subset. 

For ETH3D-SfM, the sparse depth is from COLMAP. As shown in \cref{tab:results_real_pattern}, \methodname\ significantly outperforms all baselines on the ETH3D outdoor split (RMSE=1.069 vs 1.883 for Marigold, a 43\% reduction). On the indoor split, our method works better than all DC baselines.

\subsection{Results on Synthetic Depth Patterns}

Results are shown in \cref{tab:results_virtual_pattern}. We divide the RMSE on outdoor scenes by $5.0$ to make the scale approximately match with indoors, and we report separated numbers in Appendix \cref{sec:appendix_acc_breakdown}. \methodname\ outperforms all baselines by a large margin on all depth patterns: it continues to work well with extremely sparse points (0.03\%, REL=0.034 vs 0.068 for DAv2~\cite{depth_anything_v2}), a large proportion of noise (10\% noise, RMSE=0.147 vs 0.214 for G2-MD~\cite{g2-monodepth}), or when the sparse depth map comes from SfM/sensors (ORB, RMSE=0.247 vs 0.427 for G2-MD~\cite{g2-monodepth}; LiDAR-8, RMSE=0.231 vs 0.306 for G2-MD~\cite{g2-monodepth}). These results show the superior robustness of our model across various densities, noise levels, and sensor types.

\begin{figure*}
    \vspace{-8mm}
    \centering
    \includegraphics[width=\linewidth]{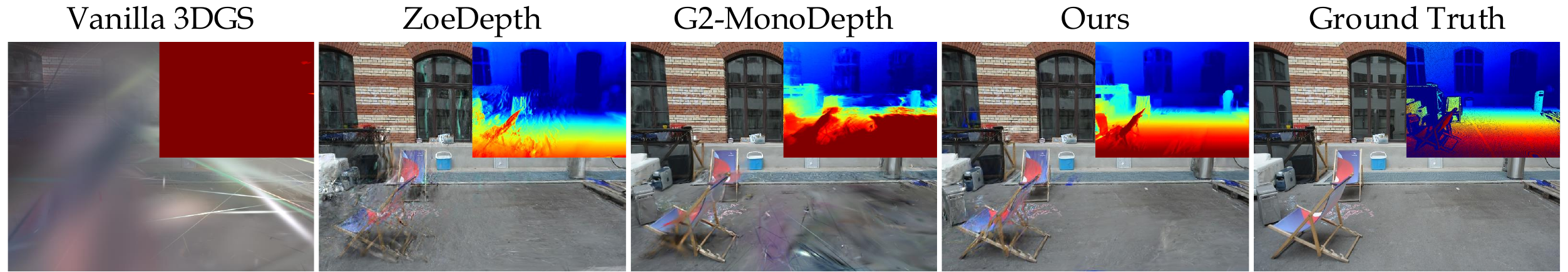}
    \vspace{-6mm}
    \caption{Rendered images and depths on test views. We train 3DGS~\cite{3dgs} with a depth loss against depth predicted by different models.}
    \vspace{-3mm}
    \label{fig:view_syn_vis_short}
\end{figure*}

\subsection{Ablation Studies}

We randomly pick an indoor and an outdoor scene from the ETH3D-SfM dataset for validation (no overlap with the test set). We also sample 1000 images from the KITTIDC training set for ablation studies (as we don't use it for training).

For the DDI ablation, we use the full training schedule, as the effectiveness is most obvious when models fully converge. Models for all other experiments are trained on $1/10$ amount of the full data due to resource constraints .

Results are shown in \cref{tab:ablation}. 1) \textit{Multi-res DDI}: the improvement is most obvious on ETH3D, where the sparse depth maps contain large holes. When using 3 resolutions, the RMSE reduces to 0.459 from 0.595 for the vanilla DDI. The metrics on KITTI are also improved. 2) \textit{ Losses}: while both the Laplacian loss and the gradient-matching loss lead to improvements over $L_1$ alone, combining them yields the best performance. 3) \textit{Depth Normalization}: using the log-depth alone sacrifices the accuracy on KITTI, as the space on the numerical axis for larger depth values is compressed. Using the log-depth plus our normalization leads to improvements on both datasets. 4) \textit{Training Depth Patterns}: using the synthetic patterns (SIFT, LiDAR) is better than training with only random samples, and injecting noise during training further boosts the performance.

\subsection{Application: Novel View Synthesis}

We show a practical application of \methodname\ on view synthesis. We run \methodname\ on the sparse depth map from COLMAP, and follow DN-Splatter~\cite{dn-splatter} to regularize the 3DGS with an additional depth loss. We test on the 13 scenes from ETH3D, and split $1/8$ of the views for validation following the standard practice. The large scale of the scenes and the small overlaps among views make it a challenging dataset for view synthesis. Further details are provided in Appendix \cref{sec:appendix_nvs}.

\begin{table}[h]
\setlength{\tabcolsep}{1.5mm}
\small
\centering
\vspace{-2mm}
        \caption{The novel view synthesis metrics and the rendered depth accuracy averaged on the 13 scenes from ETH3D.}
        \vspace{-2.5mm}
        \label{tab:view_syn_results}
            \begin{tabular}{@{}lcccc@{}}
              \hline
              \noalign{\vskip 0.2mm}  

  Methods & 3DGS & ZoeDepth~\cite{ZoeDepth} & G2-MD~\cite{g2-monodepth} & Ours \\
  \hline \hline  
  PSNR $\uparrow$ & 15.64 & 18.96 & 19.36 & \textBF{20.38} \\
  SSIM~\cite{ssim} $\uparrow$ & 0.557 & 0.573 & 0.641 & \textBF{0.660} \\
  LPIPS~\cite{lpips} $\downarrow$ & 0.418 & 0.324 & 0.273 & \textBF{0.229} \\
  RMSE (Depth) $\downarrow$ & 3.857 & 2.163 & 1.904 & \textBF{0.838}  \\
  \hline
          \vspace{-6mm}
  \end{tabular}
\end{table}

Results are shown in \cref{tab:view_syn_results} and \cref{fig:view_syn_vis_short}. The rendering and depth quality greatly improves compared to raw 3DGS, or using ZoeDepth or G2-MD for depth supervision.

\vspace{-1mm}
\section{Conclusion}
\vspace{-1mm}
We have introduced \methodname, a highly robust depth completion model that performs well on different datasets and sparse patterns, with novel designs spanning model architecture, loss, and training data. We hope \methodname\ will serve as a plug-and-play model for downstream tasks.

\clearpage

\section*{Acknowledgments}
This work was partially supported by the National Science Foundation. We thank Jing Wen and all Princeton Vision \& Learning Lab members for their insightful discussions and detailed comments on the manuscript.

{
    \small
    \bibliographystyle{ieeenat_fullname}
    \bibliography{main}
}

\clearpage

\twocolumn[
        \centering
        \Large
        \textbf{OMNI-DC: Highly Robust Depth Completion with Multiresolution Depth Integration}\\
        \vspace{1.0em}Appendix\\
        \vspace{2.0em}
       ] 

\setcounter{section}{0}
\renewcommand{\thesection}{\Alph{section}}
\setcounter{table}{0}
\renewcommand{\thetable}{\alph{table}}
\setcounter{figure}{0}
\renewcommand{\thefigure}{\alph{figure}}

\section{Comparison with Additional Generalizable DC Baselines}

The codes are not available for some of the generalizable DC baselines~\cite{SpAgNet,UniDC}, so we are only able to compare against them on NYU and KITTI. 

While all termed ``generalizable'', previous works focus on more restricted settings (TTADC and UniDC on label-free / few-shot domain adaptation; SpAgNet and DepthPrompting on generalization across sparsity), in contrast to the most challenging zero-shot, sensor-agnostic setting of our paper. As shown in \cref{tab:comparision_generalizable_baselines}, ours (zero-shot) works even better than UniDC tested on the easier 100-shot setting. On NYU, ours even outperform fully-supervised SpAgNet.

\begin{table}[ht]
\setlength{\tabcolsep}{1mm}
\footnotesize
\centering
        \caption{Numbers are copied from original papers when possible. Metric is RMSE. Ours works best under the generalizable settings (\ie, TTA/100-shot/zero-shot) on both datasets and across densities (64Lines \& 8Lines on KITTI).}
        \vspace{-2mm}
\label{tab:comparision_generalizable_baselines}
            \begin{tabular}{@{}lcccc@{}}
              \hline
  Methods & Setting & NYU-500P & KITTI-64L & KITTI-8L \\
  
  \hline \hline  
  DPrompting~\cite{depth_prompting} & \textcolor{gray}{\multirow{3}{*}{{\begin{tabular}{@{}c@{}}{Fully} \\ {Supervised} \end{tabular}}}} & \textcolor{gray}{0.105} & \textcolor{gray}{1.086} & \textcolor{gray}{1.642} \\
  SpAgNet~\cite{SpAgNet} & & \textcolor{gray}{0.114} & \textcolor{gray}{0.845} & \textcolor{gray}{2.691} \\
VPP4DC~\cite{VPP4DC} & & \textcolor{gray}{0.117} & \textcolor{gray}{0.099} & \textcolor{gray}{-} \\
  
  \hline
  TTADC~\cite{test_time_adaptation} & Test-Time Adapt. & 0.204 & - & - \\
  \hline
  UniDC~\cite{UniDC} &  \multirow{2}{*}{100-Shot} & 0.147 & 1.224 & 2.890 \\
  Drompting~\cite{depth_prompting} & & 0.175 & 1.275 & 4.587 \\
  \hline
       UniDC~\cite{UniDC} & \multirow{3}{*}{\textbf{Zero-Shot}} & 0.323 & 4.061 & - \\
       VPP4DC~\cite{VPP4DC} & & 0.247 & 1.609 & - \\
       \textbf{Ours} & & \textBF{0.111} & \textBF{1.191} & \textBF{2.058} \\
    \hline
  \end{tabular}
  \vspace{-2mm}
\end{table}

\section{Ablation Studies on Training Data}
\label{sec:appendix_data_ablations}

We show two things here: 1) Mixing real-world data for training harms performance, both qualitatively and quantitatively. 2) When using the same training datasets, our method still works better than baselines (\ie, OGNIDC~\cite{ognidc} and CompletionFormer~\cite{completionformer}). 

\begin{figure}[ht]
\vspace{-0mm}
    \centering
    \includegraphics[width=\linewidth]{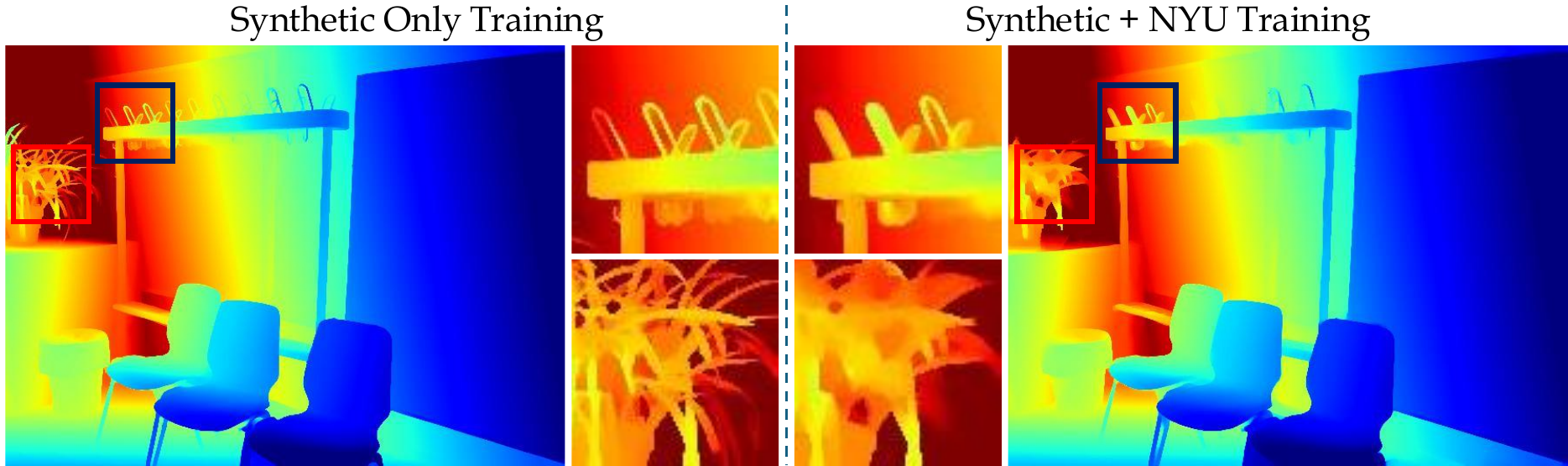}
    \vspace{-5mm}
    \caption{Mixing NYU for training produces blurry depth maps on iBims~\cite{ibims}.}
    \label{fig:synthetic_is_sharper}
\vspace{-3mm}
\end{figure}

We train OMNI-DC and baselines on either fully synthetic data, or synthetic+NYUv2. As shown in \cref{fig:synthetic_is_sharper}, synthetic+real training produces blurry results, as NYU labels from Kinect are blurry. Compared to fully synthetic training, mixing NYU for training results in worse RMSE for zero-shot testing on most (6/8) of the datasets, as shown in \cref{tab:synthetic_is_better}. Nevertheless, ours is still better than baselines when trained on synthetic+real (RMSE reduced by 29.2\% from OGNI-DC and by 36.1\% from CFormer on KITTI).

\section{Results on Radar Depth Completion}
\label{sec:radar}

\begin{figure}[ht]
    \centering
    \includegraphics[width=\linewidth]{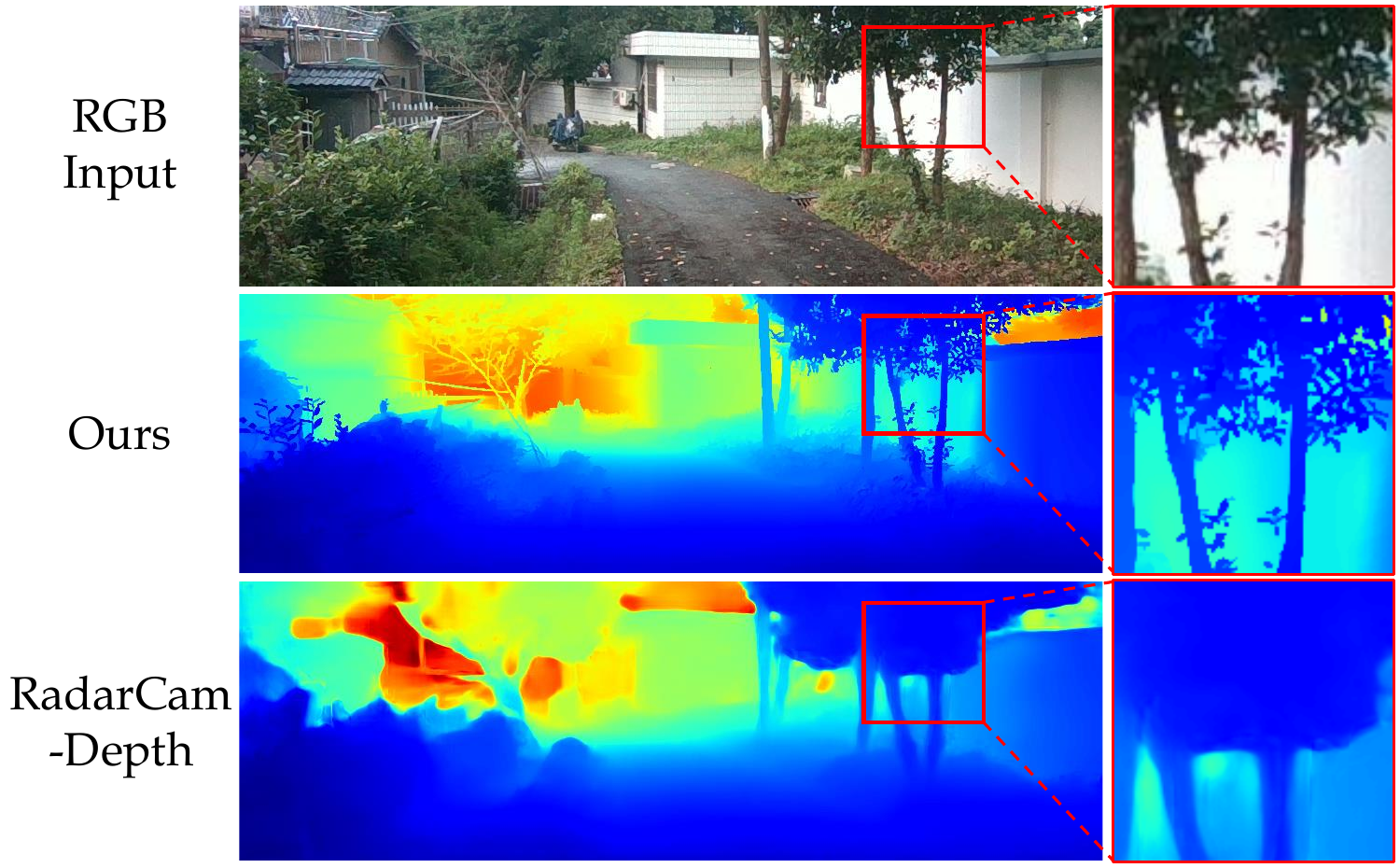}
    \caption{Results on the Radar depth completion task. Our depth maps are much sharper than RadarCam-Depth~\cite{zjuradarcam}.}
    \label{fig:zju_radar_vis}
\end{figure}

To show that OMNI-DC can generalize beyond the sparse depth patterns that it was trained on, we evaluate on the Radar-Camera fusion benchmark, ZJU-4DRadarCam~\cite{zjuradarcam}. As shown in \cref{tab:result_zju_radar}, ours outperforms all zero-shot baselines. While baselines such as G2-MD also claim to be generalizable, they perform much worse. As shown in \cref{fig:zju_radar_vis}, while our metrics slightly fall behind RadarCam-Depth (which is trained in-domain and not zero-shot), our depth map is much sharper. Sharpness is crucial for novel view synthesis applications to avoid boundary artifacts.

\begin{table}[!h]
\setlength{\tabcolsep}{0.9mm}
\footnotesize
\centering
        \caption{We follow \cite{zjuradarcam} and test with three ranges. Numbers in \textcolor{gray}{gray} are trained on ZJU-4DRadarCam and are not zero-shot; others are zero-shot. Ours outperforms all other zero-shot methods, though it falls behind methods trained in-domain.}
        \label{tab:result_zju_radar}
            \begin{tabular}{@{}lP{0.9cm}P{0.9cm}P{0.9cm}P{0.9cm}P{0.9cm}P{0.9cm}@{}}
              \hline
   \multirow{2}{*}{GT-ranges} &   \multicolumn{2}{c}{50m} &
  \multicolumn{2}{c}{70m} & \multicolumn{2}{c}{80m} \vspace{-0.7mm} \\
  \cmidrule(lr{0.5em}){2-3} \cmidrule(lr{0.5em}){4-5} \cmidrule(lr{0.5em}){6-7}
  &
  RMSE$\downarrow$ & iRMSE$\downarrow$ &
  RMSE$\downarrow$ & iRMSE$\downarrow$ &
  RMSE$\downarrow$ & iRMSE$\downarrow$  \\
  \hline \hline  
    DORN~\cite{radar_dorn} & \textcolor{gray}{4129.7} & \textcolor{gray}{31.853} & \textcolor{gray}{4625.2} & \textcolor{gray}{31.877} & \textcolor{gray}{4760.0} & \textcolor{gray}{31.879} \\
    Singh \etal~\cite{radar_singh} & \textcolor{gray}{3704.6} & \textcolor{gray}{35.342} & \textcolor{gray}{4137.1} & \textcolor{gray}{35.166} & \textcolor{gray}{4309.3} & \textcolor{gray}{35.133} \\
    RadarCam~\cite{zjuradarcam} & \textcolor{gray}{2817.4} & \textcolor{gray}{22.936} & \textcolor{gray}{3117.7} & \textcolor{gray}{22.853} & \textcolor{gray}{3229.0} & \textcolor{gray}{22.838} \\
    DA-v2~\cite{depth_anything_v2} & 5466.5 & 47.446 & 6261.3 & 47.118 & 6566.9 & 47.053 \\
    OGNI-DC~\cite{ognidc} & 7612.7 
 & 29107.2 & 8151.2 & 28800.5 & 8356.9 & 28739.0 \\
    G2-MD~\cite{g2-monodepth} & 7237.2 & 61.285 & 7980.3 & 60.803 & 8232.3 & 60.717 \\
  \hline
  \textbf{Ours} & \textbf{5256.8} & \textbf{41.477} & \textbf{5984.1} & \textbf{41.253} & \textbf{6249.1} & \textbf{41.207} \\
  \hline
  \end{tabular}
\end{table}

\begin{table*}[t]
\setlength{\tabcolsep}{0.8mm}
\footnotesize
\centering
\vspace{-3mm}
        \caption{Ablation studies on the effect of mixing real-world dataset for training. NYU consists of 1/6 of all data. All models are trained with 1/10 of the full training steps, due to resource constraints. The metric is RMSE, and the sparse depth has 0.7\% density except for NYU, VOID, and KITTI. Mixing real training data has a negative effect on most of the datasets, especially obvious outdoor. Ours works better than OGNI-DC and CFormer under both training settings.}
        \label{tab:synthetic_is_better}
            \begin{tabular}{@{}lcccccccccc@{}}
              \hline
   \multirow{2}{*}{} & \multicolumn{1}{c}{In-Domain} & \multicolumn{5}{c}{Zero-Shot, Indoors} &
  \multicolumn{3}{c}{Zero-Shot, Outdoors} &  \\
  \cmidrule(lr{1em}){2-2} \cmidrule(lr{1em}){3-7} \cmidrule(lr{1em}){8-10}
  Training &
  NYU-500P & iBims & ETH3D(In) & DIODE(In) & ARKitScenes & VOID-1500P & ETH3D(Out) & DIODE(Out) & KITTI-64L \\

  \hline \hline  
  OMNI-DC, Synthetic Only (\textBF{Ours}) & 0.119 & \textBF{0.156} & \textBF{0.118} & \textBF{0.056} & 0.023 & \textBF{0.565} & \textBF{0.322} & \textBF{2.307} & \textBF{1.279} \\
  OMNI-DC, Synthetic + NYU & \textBF{0.110} & \textBF{0.156} & 0.119 & 0.058 & \textBF{0.022} & 0.567 & 0.324 & 2.337 & 1.309 \\
  OGNI-DC~\cite{ognidc}, Synthetic Only & 0.125 & 0.164 & 0.124 & 0.063 & 0.024 & 0.573 & 0.333 & 2.332 & 1.846 \\
  OGNI-DC~\cite{ognidc}, Synthetic + NYU & 0.120 & 0.166 & 0.127 & 0.064 & 0.023 & 0.595 & 0.337 & 2.411 & 1.850 \\
  CFormer~\cite{completionformer}, Synthetic Only & 0.130 & 0.176 & 0.148 & 0.064 & 0.030 & 0.595 & 0.359 & 2.338 & 2.037 \\
  CFormer~\cite{completionformer}, Synthetic + NYU & 0.128 & 0.173 & 0.148 & 0.066 & 0.025 & 0.627 & 0.388 & 2.382 & 2.047 \\
  \hline
  \end{tabular}
\end{table*}

\begin{table*}[ht]
\setlength{\tabcolsep}{1.0mm}
\footnotesize
\centering
        \caption{Our method is robust under challenging imaging conditions (\eg, nighttime and different weathers).}

        \label{tab:robustness}
            \begin{tabular}{@{}lP{1.1cm}P{1.1cm}P{1.1cm}P{1.1cm}P{1.1cm}P{1.1cm}P{1.1cm}P{1.1cm}P{1.1cm}P{1.1cm}P{1.1cm}P{1.1cm}@{}}
              \hline 
  \multirow{2}{*}{Datasets} &
  \multicolumn{4}{c}{Carla-Night-DC~\cite{carla_nightdc}} &
  \multicolumn{2}{c}{DS-Sunny~\cite{drivingstereo}} &
  \multicolumn{2}{c}{DS-Rainy~\cite{drivingstereo}} &
  \multicolumn{2}{c}{DS-Foggy~\cite{drivingstereo}} & 
  \multicolumn{2}{c}{DS-Cloudy~\cite{drivingstereo}}
  \\
  \cmidrule(lr){2-5} \cmidrule(lr){6-7} \cmidrule(lr){8-9} \cmidrule(lr){10-11} \cmidrule(lr){12-13}   
   &
  RMSE$\downarrow$ & MAE$\downarrow$ &
  iRMSE$\downarrow$ & iMAE$\downarrow$ &
  RMSE$\downarrow$ & MAE$\downarrow$ & 
  RMSE$\downarrow$ & MAE$\downarrow$ &
  RMSE$\downarrow$ & MAE$\downarrow$ &
  RMSE$\downarrow$ & MAE$\downarrow$
  \\
  \hline \hline  
  \noalign{\vskip 0.5mm}
  
  LDCNet~\cite{carla_nightdc} & \textcolor{gray}{7.214} & \textcolor{gray}{2.014} & \underline{\textcolor{gray}{0.0546}} & \underline{\textcolor{gray}{0.0156}} & - & - & - & - & - & - & - & - \\
  DA-v2~\cite{depth_anything_v2} & 104.878 & 68.242 & 0.1560 & 0.0976 & 7.544 & 2.941 & 7.567 & 3.805 & 7.868 & 2.927 & 8.252 & 2.964 \\
  OGNI-DC~\cite{ognidc} & 13.576 & 5.469 & 0.2191 & 0.0738 & 3.774 & 
 1.494 & 5.730 & 2.384 & 3.756 & 1.654 & 3.903 & 1.499 \\
  G2-MD~\cite{g2-monodepth} & 10.488 & 3.291	& 0.0930 & 0.0246 & 3.013 & 0.875 & 2.809 & 0.982 & 3.130 & 1.149 & 3.053 & 0.872 \\

  \hline
  \textbf{Ours} & \textbf{10.068} & \textbf{2.523} & \textbf{0.0413} & \textbf{0.0105} & \textbf{2.765} & \textbf{0.741} & \textbf{2.645} & \textbf{0.844} & \textbf{2.744} & \textbf{0.909} & \textbf{2.735} & \textbf{0.714} \\
  \hline
  \end{tabular}
\end{table*}

\section{Robustness to Night Time and Bad Weather}

 As shown in \cref{tab:robustness}, our method is robust.
 
Carla-Night-DC~\cite{carla_nightdc} contains night-time driving scenes. LDCNet~\cite{carla_nightdc} is trained on Carla-Night-DC and other methods are tested zero-shot. Our method works the best, even outperforming LDCNet on iRMSE and iMAE despite never being trained specifically on night scenes. 

The DrivingStereo (DS)~\cite{drivingstereo} dataset consists of real images from driving scenes captured at different weathers. We randomly sample 500 points from GT as sparse depth. Our method consistently outperforms baselines under all weather conditions, and is more robust (OGNI-DC's RMSE $\uparrow$ 52\% under ``Rainy'' than ``Sunny'', while ours' RMSE $\downarrow$ 4\%.)

\section{Details on Novel View Synthesis}
\label{sec:appendix_nvs}

In the main paper, we have shown a practical downstream application of \methodname\ on novel view synthesis. Training neural rendering frameworks such as NeRF~\cite{nerf} or 3DGS~\cite{3dgs} on sparse input views is a challenging task, and introducing geometric priors such as depth as a regularization has been shown helpful in previous works~\cite{ds-nerf,ds-gs}. We follow the recent work DN-Splatter~\cite{dn-splatter}, and use a depth loss to train 3DGS. The loss can be written as:

\begin{equation}
    \mathcal{L} = \mathcal{L}_{\hat{C}} + 0.2 \cdot \mathcal{L}_{\hat{D}},
\end{equation}
where $\mathcal{L}_{\hat{C}}$ is the original photometric loss in 3DGS~\cite{3dgs}, and $\mathcal{L}_{\hat{D}}$ is the edge-aware depth loss proposed in \cite{dn-splatter}.

We evaluate on the ETH3D~\cite{eth3d} dataset with 13 scenes, each containing 14-76 images. The scales of the scenes are large, creating a challenging sparse view setting. We compare against the vanilla 3DGS with no depth supervision, as well as supervising with the depth map obtained from the monocular depth model ZoeDepth~\cite{ZoeDepth}, and the depth completion model G2-MD~\cite{g2-monodepth}. For ZoeDepth, we align the scale and shift against the COLMAP sparse depth, following DN-Splatter~\cite{dn-splatter}. For G2-MD and our method, we run depth completion on the COLMAP sparse depth. In addition to the results presented in the paper, we also compare against the state-of-the-art multi-view stereo (MVS) method, MVSFormer++~\cite{mvsformerpp}.

We randomly split $1/8$ of the view as test views and use the rest for training. The training follows the \cite{dn-splatter} schedule for 30K steps. We have reported the image quality statistics PSNR, SSIM, and LPIPS, as well as the RMSE between the rendered depth and the ground-truth depth on test views.

\begin{table}[ht]
\setlength{\tabcolsep}{0.9mm}
\centering
        \caption{The novel view synthesis metrics and the depth accuracy averaged on the 13 scenes from ETH3D.}
        \label{tab:view_syn_results_appendix}
            \begin{tabular}{@{}lccccc@{}}
              \hline
              \noalign{\vskip 0.5mm}  
  \noalign{\vskip 0.5mm}

  Methods & 3DGS & \begin{tabular}{@{}c@{}}{Zoe-} \\ {Depth} \end{tabular} & G2-MD & \begin{tabular}{@{}c@{}}{MVS-} \\ {Former++} \end{tabular} & Ours \\
  \hline \hline  
  PSNR $\uparrow$ & 15.64 & 18.96 & 19.36 & 20.02 & \textBF{20.38} \\
  SSIM $\uparrow$ & 0.557 & 0.573 & 0.641 & 0.644 & \textBF{0.660} \\
  LPIPS $\downarrow$ & 0.418 & 0.324 & 0.273 & 0.254 & \textBF{0.229} \\
  RMSE (Depth) $\downarrow$ & 3.857 & 2.163 & 1.904 & 1.847 & \textBF{0.838}  \\

  \hline
  \end{tabular}
\end{table}

As shown in \cref{tab:view_syn_results_appendix}, \methodname\ outperforms all methods in terms of both rendering and geometry reconstruction quality. 

\begin{figure*}
    \centering
        \vspace{-5mm}
    \includegraphics[width=\linewidth]{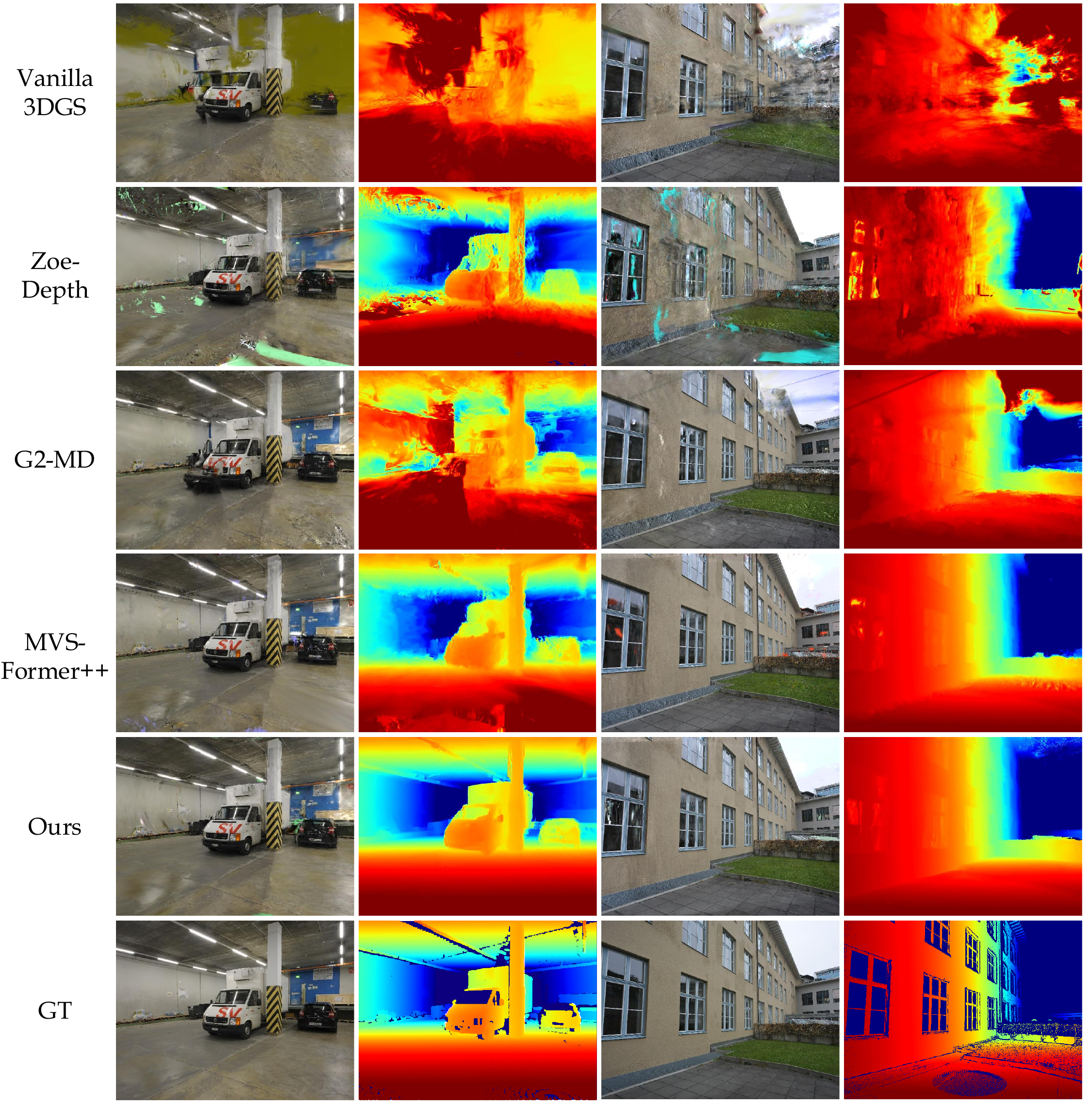}
     \vspace{-5mm}
    \caption{Visualization of the rendered images and rendered depth maps against ground-truth on test views of the ETH3D dataset. The vanilla 3DGS is trained with only the photometric loss, and all other rows are trained with a depth loss against the predicted depth maps of the corresponding models. Our model generates significantly higher quality images and geometry (depth maps).}
    \label{fig:view_syn_vis}
    \vspace{-3mm}    
\end{figure*}

More visualizations are shown in \cref{fig:view_syn_vis}. The 3DGS regularized with our depth maps produces much fewer floater artifacts compared to baselines. This shows that users can directly use our \methodname\ to improve the 3DGS quality, without any retraining for the depth model.

\begin{table*}[t]
\setlength{\tabcolsep}{1.12mm}
\vspace{-5mm}
\centering
        \caption{Results on the NYUv2 dataset with 5-500 random samples. The numbers in \textcolor{gray}{gray} are trained on NYU with 500 points, and we exclude them from the ranking. On relatively dense inputs, our method works the best among all the methods tested zero-shot, and is very close to the best model trained on NYU (REL=0.014 vs 0.011 for DFU~\cite{DFU} on NYU-500). On NYU-5, our method works better than all DC baselines (RMSE=0.536 vs 0.633 for OGNI-DC~\cite{ognidc}).}
        \vspace{-2mm}
        \label{tab:nyu_results}
            \begin{tabular}{@{}lP{2.2cm}P{1.12cm}P{1.12cm}P{1.12cm}P{1.12cm}P{1.12cm}P{1.12cm}P{1.12cm}P{1.12cm}P{1.12cm}P{1.12cm}@{}}
              \hline
              \noalign{\vskip 0.5mm}  
  \noalign{\vskip 0.5mm}
  & \multirow{2}{*}{Methods} &
  \multicolumn{2}{c}{NYU-500} &
  \multicolumn{2}{c}{NYU-200} & 
  \multicolumn{2}{c}{NYU-100} &
  \multicolumn{2}{c}{NYU-50} & 
  \multicolumn{2}{c}{NYU-5}
  \\
  \cmidrule(lr){3-4} \cmidrule(lr){5-6}
  \cmidrule(lr){7-8}
  \cmidrule(lr){9-10}
  \cmidrule(lr){11-12}
  & &
  RMSE & REL &
  RMSE & REL &
  RMSE & REL & 
  RMSE & REL & 
  RMSE & REL 
  \\
  \hline \hline  
  \noalign{\vskip 0.5mm}
   \multirow{5}{*}{\begin{tabular}{@{}c@{}}{Trained on} \\ {NYU} \end{tabular}} & CFormer~\cite{completionformer} & \textcolor{gray}{0.090} & \textcolor{gray}{0.012} &
 \textcolor{gray}{0.141} &  \textcolor{gray}{0.021} & \textcolor{gray}{0.429} & \textcolor{gray}{0.092} & \textcolor{gray}{0.707} & \textcolor{gray}{0.181} & \textcolor{gray}{1.141} & \textcolor{gray}{0.307} \\
  & DFU~\cite{DFU} & \textcolor{gray}{0.091} & \textcolor{gray}{0.011} & - & - & - & - & - & - & - & - \\
  & BP-Net~\cite{BP-Net} & \textcolor{gray}{0.089} & \textcolor{gray}{0.012} &  \textcolor{gray}{0.132} &  \textcolor{gray}{0.021} & \textcolor{gray}{0.414} & \textcolor{gray}{0.090} & \textcolor{gray}{0.609} & \textcolor{gray}{0.157} & \textcolor{gray}{0.869} & \textcolor{gray}{0.294} \\
 & DPromting~\cite{depth_prompting} & \textcolor{gray}{0.105} & \textcolor{gray}{0.015} & \textcolor{gray}{0.144} & \textcolor{gray}{0.023} & \textcolor{gray}{0.178} & \textcolor{gray}{0.031} & \textcolor{gray}{0.213} & \textcolor{gray}{0.043} & \textcolor{gray}{0.380} & \textcolor{gray}{0.095} \\ 
  & OGNI-DC~\cite{ognidc} & \textcolor{gray}{0.089} & \textcolor{gray}{0.012} &  \textcolor{gray}{0.124} &  \textcolor{gray}{0.018} & \textcolor{gray}{0.157} & \textcolor{gray}{0.025} & \textcolor{gray}{0.207} & \textcolor{gray}{0.038} & \textcolor{gray}{0.633} & \textcolor{gray}{0.171} \\
  \hline
  \multirow{5}{*}{Zero-shot} & Depth Pro~\cite{depth_pro} & \cellcolor{yellow!50} 0.266 & 0.062 & \cellcolor{yellow!50} 0.266 & 0.062 & \cellcolor{yellow!50} 0.266 & \cellcolor{yellow!50} 0.062 & \cellcolor{orange!50}  0.266 & \cellcolor{orange!50} 0.062 & \cellcolor{red!50} 0.266 & \cellcolor{red!50} 0.062 \\
  & DA-v2~\cite{depth_anything_v2} & 0.309 & \cellcolor{yellow!50} 0.061 & 0.309 & \cellcolor{yellow!50} 0.061 & 0.314 & \cellcolor{yellow!50} 0.062 & 0.330 & 0.063 & 0.814 & \cellcolor{orange!50} 0.136 \\
  & Marigold~\cite{marigold} & 0.426 & 0.115 & 0.428 & 0.116 & 0.431 & 0.117 & 0.436 & 0.118 & \cellcolor{yellow!50}  0.545 & 0.150 \\
  & G2-MD~\cite{g2-monodepth} & \cellcolor{orange!50} 0.122 & \cellcolor{orange!50} 0.017 & \cellcolor{orange!50} 0.169 & \cellcolor{orange!50} 0.027 & \cellcolor{orange!50} 0.222 & \cellcolor{orange!50} 0.038 & \cellcolor{yellow!50} 0.286 & \cellcolor{yellow!50} 0.056 & 0.744 & 0.207 \\

  \hhline{>{\arrayrulecolor [gray]{1.0}}->{\arrayrulecolor {black}}-----------}
  & \textbf{Ours} & \cellcolor{red!50} 0.111 & \cellcolor{red!50} 0.014 & \cellcolor{red!50} 0.147 & \cellcolor{red!50} 0.021 & \cellcolor{red!50} 0.180 & \cellcolor{red!50} 0.029 & \cellcolor{red!50} 0.225 & \cellcolor{red!50} 0.041 & \cellcolor{orange!50} 0.536 & \cellcolor{yellow!50}  0.142  \\
  \hline
  \end{tabular}
  \vspace{-3mm}
\end{table*}

\section{Implementation Details}
\label{sec:appendix_implementation_details}

\subsection{Model Architecture and Loss Functions}
We use the CompletionFormer~\cite{completionformer} as the backbone. CompletionFormer is a U-Net-like~\cite{UNet} architecture with a feature pyramid. We extract the depth gradients by using the $1/4$ resolution feature map with a series of ResNet~\cite{ResNet} blocks and $\operatorname{MaxPool2D}$ layers, to obtain the depth gradients at the $1/4$, $1/8$, and $1/16$ resolution. 

From the full-resolution feature map, we extract the parameters for the DySPN~\cite{DySPN} (propagation weights and confidence) and scale parameters for computing the Laplacian loss. Specifically, since the scale parameter $b$ must be positive, we parameterize it as $b = \exp(\gamma)$ following \cite{SEA_RAFT}, and predict $\gamma$ from a $\operatorname{Conv}$ layer. We clamp the minimum value of $\gamma$ to $-2.0$ to stabilize training.

To better deal with the noise in the input depth, we follow OGNI-DC~\cite{ognidc} and use a $\operatorname{sigmoid}$ layer to predict a confidence map for the input sparse depth.  Denote the confidence map as $\mathbf{\hat{C}} \in (0,1)^{H \times W}$, the sparse depth energy term is re-weighted as (see Eqn.3 in the main paper):

\begin{equation}
        \vspace{-2mm}
\label{eqn:ognidc-energy-confidence}
    \mathcal{E}_O = \sum_{i,j}^{W,H} \mathbf{M}_{i,j} \cdot \mathbf{C}_{i,j} \cdot (\mathbf{D}_{i,j} - \mathbf{O}_{i,j})^2
\end{equation}

When $C_{i,j} \rightarrow 0$, the contribution of the corresponding sparse depth point becomes zero, providing a data-driven mechanism for the network to ignore the noisy depths. Unlike OGNI-DC which trains the confidence map through the depth loss, we record the noisy pixels when generating the virtual sparse pattern and use an axillary binary cross-entropy loss to directly supervise the confidence map.

The gradient-matching loss is implemented following MegaDepth~\cite{midas} and MiDaS~\cite{midas}:

\begin{equation}
\label{eqn:gm_loss}    
   \mathcal{L}_{\text{gm}} = \frac{1}{HW} \sum_{k=1}^{4} \sum_{i,j}^{W,H} \left( \left| \nabla_x R_{i,j}^k \right| + \left| \nabla_y R_{i,j}^k \right| \right),
\end{equation}

Where $R^1 = \mathbf{\hat{D}} - \mathbf{D}^{\text{gt}}$. Similarly, $R^k$ is the depth difference at the $k$\textsuperscript{th} resolution.

\subsection{Training Details}

The model is trained with an Adam~\cite{Adam} optimizer with an initial learning rate of $1e-3$, for a total of 72 epochs. The learning rate decays by half at the 36\textsuperscript{th}, 48\textsuperscript{th}, 56\textsuperscript{th}, and 64\textsuperscript{th} epochs, following \cite{completionformer}. 

Since the five training datasets are vastly different in size, we uniformly sample 25K images from each dataset to balance their contributions in each epoch. We also normalize the median depth values of all training samples to $1.0$ to balance the loss among different types of scenes. 

We sample the random samples, SfM keypoints, and LiDAR points with a ratio of 2:1:1. This ratio empirically yeilds good performance, but the performance of our model is not sensitive to it. Random point densities are sampled in the range $0.03\% \sim 0.65\%$ (\ie, $100 \sim 2000$ points). The SfM points are sampled at the SIFT~\cite{SIFT} keypoints. For the random and SfM points, we also inject $0\% \sim 5\%$ noisy depths by random sampling between the 5\textsuperscript{th} and 95\textsuperscript{th} percentile interval of the image depth range. When generating the LiDAR keypoints, we randomize the number of lines, the center of the LiDAR, and the camera intrinsics. We additionally synthesize the boundary error caused by the baseline between the camera and the LiDAR. Specifically, we random sample a virtual viewpoint for the LiDAR., and project the depth to the virtual view. This leaves holes in the projected depth map, so we use the heuristic-based inpainting used in LRRU~\cite{LRRU} to fill those holes. We finally sample the LiDAR points from the virtual view, and project it back to the original view.


\section{Limitations} 
\label{sec:limitations}

\begin{figure}[ht]
    \centering
    \vspace{-2mm}
    \includegraphics[width=\linewidth]
{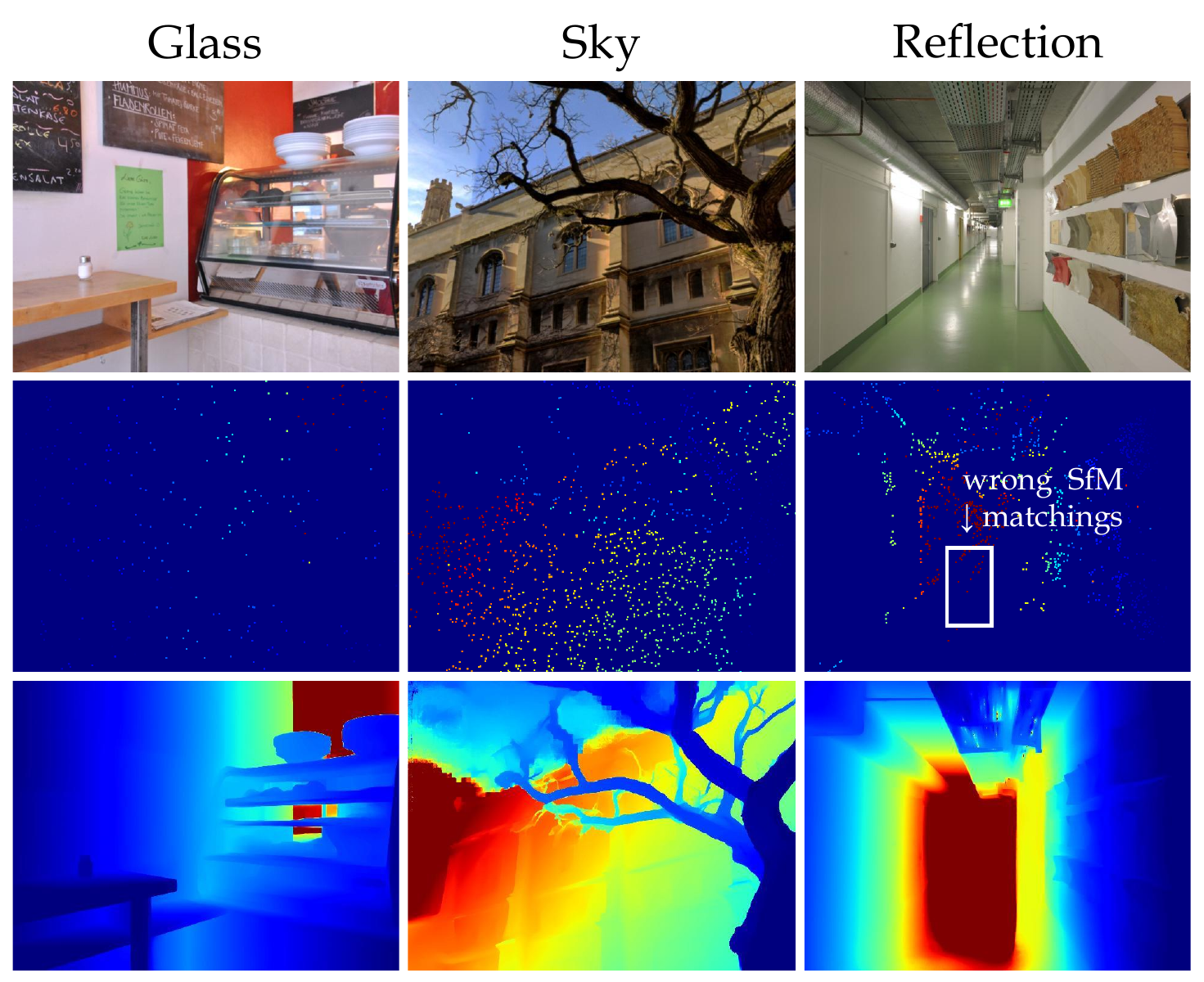}
    \vspace{-5mm}
    \caption{Failure cases of \methodname. Our model makes erroneous predictions when the scene contains glasses or reflective surfaces, as the depth sensor or multiview matching may fail. The sky cannot be naturally represented in the linear depth space.}
    \label{fig:failure_cases}
    \vspace{-5mm}
\end{figure}

\textbf{1)} Like other depth estimation models, our method faces challenges when predicting depth for transparent surfaces (\eg, glasses), reflective surfaces, or the sky. In \cref{fig:failure_cases} we show a few failure cases. \textbf{2)} The backbone of our method takes 4 channels (RGB-D) input, which makes it hard to benefit from the pre-trained models designed for RGB images, such as DINO-v2~\cite{dinov2}. One possible direction is removing the depth channel from the feature extractor. \textbf{3)} Our model currently cannot deal with the case with no sparse depth inputs (\ie, monocular depth estimation). Having the model's performance degrade more smoothly when the input depths become sparser is a future direction. \textbf{4)} The current model cannot handle certain types of sparse depth patterns very well, such as the radar inputs discussed in \cref{sec:radar}, and large holes that may appear in object removal/inpainting applications. Expanding the sparse depth synthesis pipeline to cover these during training is a promising direction.

\section{More Results on the NYUv2 Dataset}

We show results on more densities in \cref{tab:nyu_results}. We exclude all the in-domain DC baselines trained on the NYU training set from the ranking. Our method works better than all zero-shot baselines on the 500, 200, 100, and 50 densities. On the original setting of NYUv2 (NYU-500), our method has a close performance to the best model trained on NYU (REL=0.014 vs 0.011 for DFU~\cite{DFU}). On the extremely sparse case (NYU-5), our method works better than OGNI-DC~\cite{ognidc} and G2-MD~\cite{g2-monodepth}, although worse than the monocular depth methods such as Depth Pro~\cite{depth_pro}.

\begin{figure*}[t]
    \centering
    \vspace{-2mm}
\includegraphics[width=\linewidth]{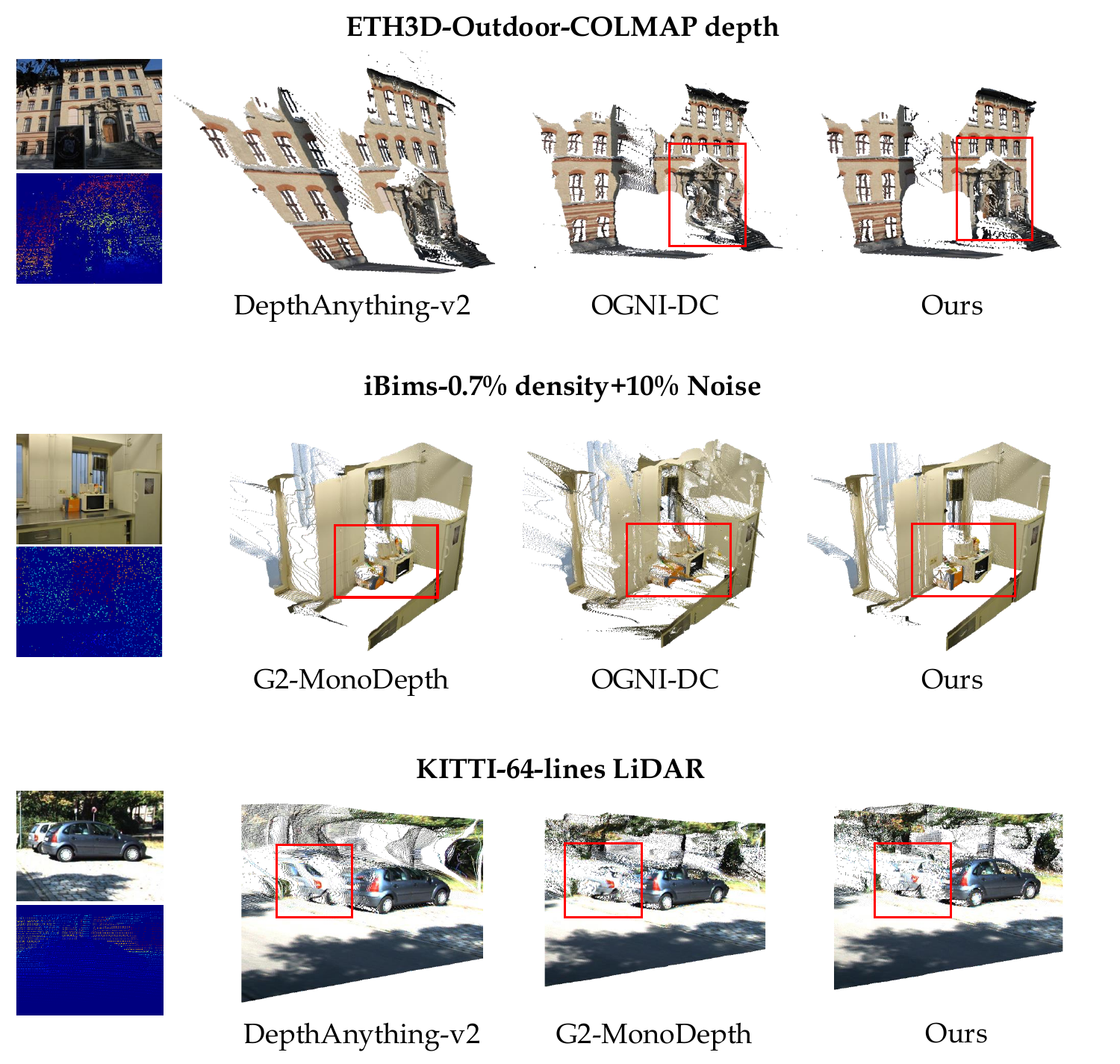}
    \caption{The qualitative comparison of the 3D structures between our method and the best-performing baselines. On the outdoor scene from ETH3D, DA-v2~\cite{depth_anything_v2} has trouble capturing the global structure, while OGNI-DC's reconstruction has distorted local details. On the noisy sparse depth map on iBims, the OGNI-DC's prediction is greatly distorted by the outliers, and our method is robust to noise. On KITTI, our method is able to reconstruct the high-quality 3D structure of the white car.}
    \label{fig:qualitative_3D}
    
\end{figure*}

\section{Visualizations of Point Cloud}
\label{sec:qualitative_3D}
We visualize the 3D reconstruction quality of our predicted depth map by projecting the depth map into 3D using the ground-truth camera intrinsics. We also compared against the few strongest baselines, \ie, DepthAnythingv2~\cite{depth_anything_v2}, OGNI-DC~\cite{ognidc}, and G2-MonoDepth~\cite{g2-monodepth}. As shown in \cref{fig:qualitative_3D}, our method achieves better results in both global structures (orientation of the walls) and local details (cars).

\section{Details on Evaluation Datasets}

\begin{figure*}
    \centering
    \vspace{-5mm}
    \includegraphics[width=\linewidth]{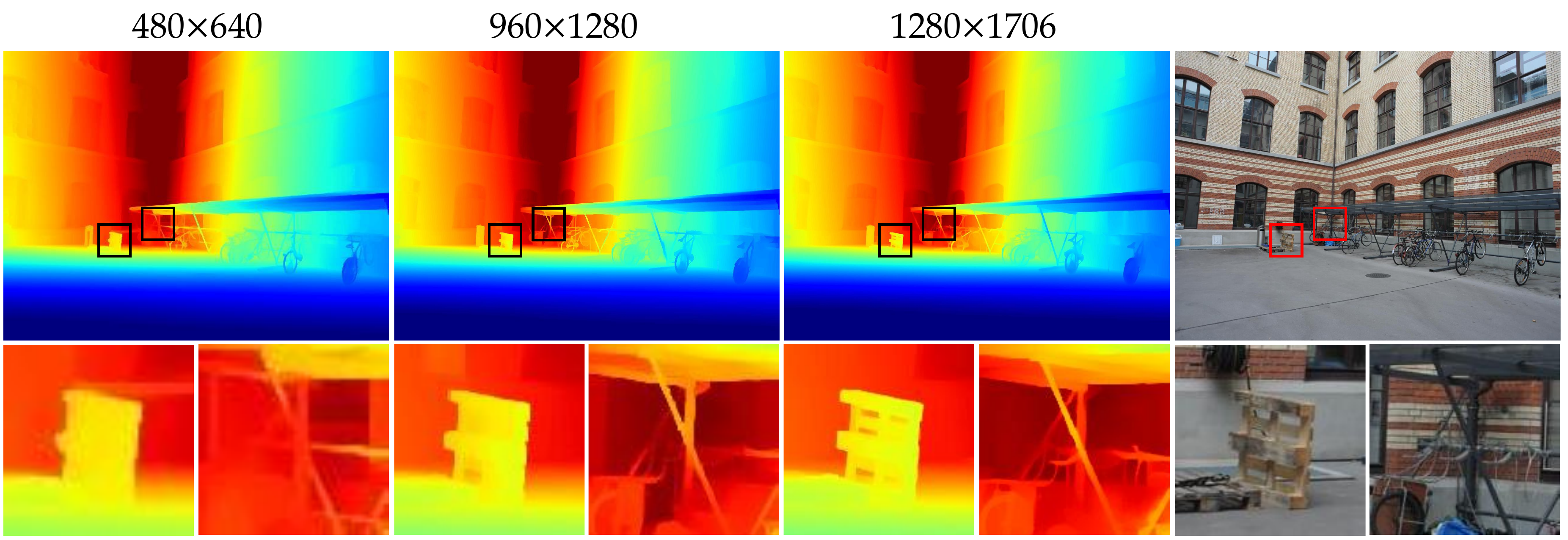}
    \caption{More details are captured when running inference with higher resolution images at test time. All sparse depths are sampled under the 0.7\% density.}
    \label{fig:resolution_scale_up}
    \vspace{-5mm}
\end{figure*}

We list the details of the datasets we use below. Samples from the datasets can be found in \cref{fig:2d_vis_p3,fig:2d_vis_p1,fig:2d_vis_p2}.

\textbf{iBims}~\cite{ibims} consists of 100 indoor scenes captured with a laser scanner. The original images are at 480$\times$640 resolution.

\textbf{ARKitScenes}~\cite{arkitscenes} is a large scale dataset consisting of more than 450K frames of scans of 5K indoor scenes. The validation split contains about 3.5K images in the landscape orientation, from which we randomly pick 800 images as our test set. The original high-res laser-scan images are at resolution 1440$\times$1920, from which we resize to 480$\times$640.

\textbf{ETH3D}~\cite{eth3d}'s test set contains 13 scenes total with 454 images, with ground-truth captured using a laser scanner. The original images are at 4032$\times$6048 resolution, from which we downsample at approximately a factor of 8 to 480$\times$640. We pick the ``office" and the ``courtyard" scene as the validation set, and further split the rest 11 scenes into indoors (6 scenes, 193 images) and outdoors (5 scenes, 197 images). For the real SfM patterns, we project the visible keypoints from the COLMAP~\cite{COLMAP_SfM} reconstruction for each scene into 2D to construct the sparse depth map.

\textbf{DIODE}~\cite{diode}'s validation split contains 3 indoor scenes and 3 outdoor scenes, with 325 and 446 images in total respectively. The ground truth is captured with a FARO laser scanner. We find that the original depth measurements at occlusion boundaries are very noisy. Therefore, we filter out the pixel whose depth is different from its neighboring pixels by more than 5\% (indoor) and 15\% (outdoor). This effectively removes the noise while preserving most of the useful information. Images are resized to 480$\times$640.

\textbf{KITTI}~\cite{KITTIDC}'s validation set contains 1000 images from 5 scenes in total. We subsample the original 64-line LiDAR by clustering the elevation angles of the LiDAR points to construct the virtual 16-line and 8-line input following \cite{twin_surface_dc}. We crop the top 96 pixels containing only sky regions, resulting in an image resolution of 256$\times$1216.

\section{Test-Time Scaling Up to Higher-Resolution Images}

Most of the experiments in this paper are conducted under the resolution of 480$\times$640. However, modern cameras can often capture images at a higher resolution, which captures more details. Therefore, it is desirable that our DC model can work under higher resolutions.

We feed \methodname\ with high-resolution images at test time. As shown in \cref{tab:resolution_scale_up}, the inference time is 2.1$\times$ and $3.6\times$ longer when tested on images with $2\times$ and $2.7\times$ resolution, respectively, a lower rate compared to the increase in pixel count. The memory consumption is 11.1GB when tested under the resolution of 1280$\times$1706, which can be held on a 12GB GPU such as an RTX 4070.

\begin{table}[ht]
\setlength{\tabcolsep}{1.4mm}
\centering
\vspace{-1mm}
        \caption{Speed ane memory consumption on higher resolutions. Numbers benchmarked on a 3090 GPU.}
        \vspace{-1mm}
        \label{tab:resolution_scale_up}
            \begin{tabular}{@{}lccc@{}}
              \hline
              \noalign{\vskip 0.5mm}  
  \noalign{\vskip 0.5mm}

  Resolution & 480$\times$640 & 960$\times$1280 & 1280$\times$1706 \\
  \hline \hline  
  Inference Time (ms) & 235 & 495 & 839 \\
  Memory (GB) & 4.6 & 7.9 & 11.1 \\
  \noalign{\vskip 0.5mm}

  \hline
  \end{tabular}
\end{table}

Qualitative results are shown in \cref{fig:resolution_scale_up}. While \methodname\ is trained on a low resolution (480$\times$640), it can generalize to higher resolution images at test time,  producing higher quality depth maps.

The results show that \methodname\ has a strong capability of scaling up to higher-resolution images at test time.

\section{Guaranteed Scale Equivariance}
\label{sec:appendix_scale_equivariance_proof}

Scale equivariance means the scale of the output depth respects the scale of the input depth. For example, when the input is given in the unit of millimeters ($mm$), the output should also be in millimeters. This is a desired property, as it makes the system simple to use. For example, if a DC model is not scale-equivariant, the user will have to convert it to metric space before feeding it into the DC model, which requires estimating the arbitrary scale factor from their COLMAP reconstruction and could be impossible. 

Assume $F$ to be a DC model taking the RGB image $\mathbf{I}$ and the sparse depth map $\mathbf{O}$ as input, and outputs a dense depth map $\mathbf{\hat{D}}$, \ie,

\begin{equation}
    \mathbf{\hat{D}} = F(\mathbf{I}, \mathbf{O}).
\end{equation}

We formally define the equivariance property as follows:
\begin{equation}
   F(\mathbf{I}, \beta \cdot \mathbf{O}) = \beta \cdot F(\mathbf{I}, \mathbf{O}), \forall \beta \in \mathbb{R}_{+},
\end{equation}
where $\beta$ is an arbitrary scale factor. For example, $\beta=1000$ when converting depth from meters ($m$) into millimeters ($mm$).

We first theoretically prove that \methodname\ is guaranteed to be scale equivariant, and then confirm it by empirical results.

\subsection{Theoretical Proof}

We first show that the input to the neural network is \textit{invariant} to the scale of the input depth. Recall that we normalize the input depth values to the neural network by its median:

\begin{equation}
    \mathbf{\hat{G}} = F(\mathbf{I}, \mathbf{\Tilde{O}}; \theta), \mathbf{\Tilde{O}} = \log(\mathbf{O}) - \log(\operatorname{median}(\mathbf{O})).
\end{equation}

It is easy to see that $\mathbf{\Tilde{O}}$ is invariant to the input scale, \ie,

\begin{equation}
\begin{split}
    & \mathbf{\Tilde{O}}(\beta \cdot \mathbf{O}) = \log(\beta \cdot\mathbf{O}) - \log(\operatorname{median}(\beta \cdot \mathbf{O})) \\
    &= \log(\beta) + \log(\mathbf{O}) - \log(\beta) - \log(\operatorname{median}(\mathbf{O})) \\
    &= \mathbf{\Tilde{O}}(\mathbf{O}), \forall \beta \in \mathbb{R}_{+}.
\end{split}
\end{equation}

Correspondingly, the output of the neural network, $\mathbf{\hat{G}}$, is also invariant to the input scale, because all its input is scale-invariant:

\begin{equation}
    \mathbf{\hat{G}}(\mathbf{I}, \beta \cdot \mathbf{O}) = \mathbf{\hat{G}}(\mathbf{I}, \mathbf{O}), \forall \beta \in \mathbb{R}_{+}.
\end{equation}

We therefore omit the input of $\mathbf{\hat{G}}$ and treat it as a constant in the following deductions.

Note that the depth integration is done in the $\log$-depth space, and recall the energy terms are:

\begin{equation}
    \label{eq:optim_target_appendix}
    \mathbf{\hat{D}}^{\log} = \argmin_{\mathbf{D}^{\log}} \left(\alpha \cdot \mathcal{E}_O (\mathbf{D}^{\log}, \mathbf{O},\mathbf{M} ) + \mathcal{E}_G (\mathbf{D}^{\log}, \mathbf{\hat{G}})  \right),
\end{equation}
where
\begin{equation}
\label{eqn:ognidc-energy_appendix}
\begin{split}
    \mathcal{E}_O &:= \sum_{i,j}^{W,H} \mathbf{M}_{i,j} \cdot (\mathbf{D}^{\log}_{i,j} - \log(\mathbf{O}_{i,j}))^2, \\
      \mathcal{E}_G &:=  \sum_{r=1}^R \sum_{i,j}^{W,H} \left( \mathbf{G}^x_{i,j} - \mathbf{\hat{G}}^x_{i,j} \right) ^ 2
    + \left(\mathbf{G}^y_{i,j} - \mathbf{\hat{G}}^y_{i,j} \right) ^ 2,
\end{split}
\end{equation}
with $\mathbf{G}^{r, x}_{i,j} := \mathbf{D}^{r}_{i,j} - \mathbf{D}^{r}_{i-1, j}$; $\mathbf{G}^{r, y}_{i,j} := \mathbf{D}^{r}_{i,j} - \mathbf{D}^{r}_{i, j-1}$ being the analytical gradients at the resolution $r$.

We write $\mathbf{\hat{D}}^{\log}$ as a function of $\mathbf{\hat{G}}$, $\mathbf{O}$, and $\mathbf{M}$, i.e., $\mathbf{\hat{D}}^{\log}(\mathbf{\hat{G}}, \mathbf{O}, \mathbf{M})$. Given the above definition, we have the lemma below:
\newtheorem{theorem}{Theorem}
\newtheorem{lemma}[theorem]{Lemma}
\begin{lemma}
\label{lemma_1}
If $\mathbf{\hat{D}}^{\log}(\mathbf{\hat{G}}, \mathbf{O}, \mathbf{M})$ is the optimal solution to \cref{eq:optim_target_appendix}, then $\log{\beta} + \mathbf{\hat{D}}^{\log}(\mathbf{\hat{G}}, \mathbf{O}, \mathbf{M})$ is the optimal solution if we multiply $\mathbf{O}$ by $\beta$, \ie, $\mathbf{\hat{D}}^{\log}(\mathbf{\hat{G}}, \beta \cdot \mathbf{O}, \mathbf{M}) = \log{\beta} + \mathbf{\hat{D}}^{\log}(\mathbf{\hat{G}}, \mathbf{O}, \mathbf{M})$, $\forall \beta \in \mathbb{R}_{+}.$
\end{lemma}

This can be seen from the linearity of \cref{eqn:ognidc-energy_appendix}. Plugging $\log{\beta} + \mathbf{D}^{\log}$ and $\beta \cdot \mathbf{O}$ into \cref{eqn:ognidc-energy_appendix} gives the exact same energy as $\mathbf{D}^{\log}$ and $\mathbf{O}$.

Given Lemma~\ref{lemma_1}, we finally have 

\begin{equation}
\begin{split}
    \mathbf{\hat{D}}(\mathbf{\hat{G}}, \mathbf{\beta \cdot O}, \mathbf{M}) &= \exp \left(\mathbf{\hat{D}}^{\log}(\mathbf{\hat{G}}, \beta \cdot \mathbf{O}, \mathbf{M})\right) \\
    &= \exp \left(\log{\beta} + \mathbf{\hat{D}}^{\log}(\mathbf{\hat{G}}, \mathbf{O}, \mathbf{M})\right) \\
    &= \beta \cdot \mathbf{\hat{D}}(\mathbf{\hat{G}}, \mathbf{O}, \mathbf{M}), \forall \beta \in \mathbb{R}_{+}. \;\square
\end{split}
\end{equation}

\subsection{Empirical Evidence}

\begin{table}[ht]
\setlength{\tabcolsep}{1.7mm}
\centering

        \caption{Guaranteed Depth Scale Equivalence. Metric is REL.}
        \label{tab:scale_equivariance}
            \begin{tabular}{@{}lccccc@{}}
              \hline
              \noalign{\vskip 0.5mm}  
  \noalign{\vskip 0.5mm}

  Depth Scale & 0.001$\times$ & 0.1$\times$ & 1$\times$ & 10$\times$ & 1000$\times$ \\
  \hline \hline  
  \noalign{\vskip 0.5mm}
  CFormer~\cite{completionformer} & 810.8 & 5.404 & 0.236 & 0.684 & 0.997 \\
  OGNI-DC~\cite{ognidc} & 7.079 & 0.704 & 0.158 & 0.387 & 0.622  \\
  G2-MD~\cite{g2-monodepth} & 0.386 & 0.187 & 0.108 & 2.693 & 145.1 \\
  \hline
  \textbf{Ours} & 0.081 & 0.081 & 0.081 & 0.081 & 0.081 \\

  \hline
  \end{tabular}
\end{table}

We test \methodname\ and several baselines on the ETH3D-SfM-Indoor validation split. In each column, we multiply both the input sparse depth and ground-truth depth by a scale factor and compute the relative error:
\begin{equation}
    \operatorname{REL}(\mathbf{\hat{D}}, \mathbf{D}^{\text{gt}}) = \frac{1}{HW} \cdot \sum_{i,j}^{W,H} \frac{|\mathbf{\hat{D}}_{i,j} - \mathbf{D}^{\text{gt}}_{i,j}|}{\mathbf{D}^{\text{gt}}_{i,j}}
\end{equation}

The REL error should be a constant across all scales if the model has the scale-equivariance property. Results are shown in \cref{tab:scale_equivariance}. Our method has the same REL error across all scales, proving the guaranteed scale equivariance in our implementation. All baselines fail catastrophically on the extreme cases (\eg, $\times 1000$ when from $m$ to $mm$).

\section{Evaluation Details}

\subsection{Baselines}

We run Depth Pro~\cite{depth_pro} to directly predict metric depth, without considering the sparse depth input. We estimate the global scale and shift in the least square manner against the sparse depth points for Marigold~\cite{marigold} (in linear depth space) and DepthAnythingv2~\cite{depth_anything_v2} (in disparity space).

For BP-Net~\cite{BP-Net}, Depth Prompting~\cite {depth_prompting}, and OGNI-DC~\cite{ognidc}, we use their model trained on NYUv2 and KITTI for indoor and outdoor testing, respectively. We use the DFU~\cite{DFU} checkpoint trained on KITTI for all experiments, since its NYU code is not released. G2-MD~\cite{g2-monodepth} needs a separate scaling factor for indoors and outdoors, and we use 20.0 and 100.0 as suggested by the authors.

Note that while we provide the most favorable settings for all baselines, our method has only a \textit{single model} and does \textit{not} need separate hyperparameters for indoor and outdoor scenes, making it the simplest to use.

\subsection{Evaluation Metrics}

The metrics are defined as follows:

\begin{equation*}
\begin{split}
    \operatorname{MAE}(\mathbf{\hat{D}}, \mathbf{D}^{\text{gt}}) &= \frac{1}{HW} \cdot \sum_{i,j}^{W,H} |\mathbf{\hat{D}}_{i,j} - \mathbf{D}^{\text{gt}}_{i,j}| \\
\operatorname{REL}(\mathbf{\hat{D}}, \mathbf{D}^{\text{gt}}) &= \frac{1}{HW} \cdot \sum_{i,j}^{W,H} \frac{|\mathbf{\hat{D}}_{i,j} - \mathbf{D}^{\text{gt}}_{i,j}|}{\mathbf{D}^{\text{gt}}_{i,j}}
\end{split}
\end{equation*}

\begin{equation*}
\begin{split}
    & \operatorname{RMSE}(\mathbf{\hat{D}}, \mathbf{D}^{\text{gt}}) = \sqrt{ \frac{1}{HW} \cdot \sum_{i,j}^{W,H} (\mathbf{\hat{D}}_{i,j} - \mathbf{D}^{\text{gt}}_{i,j})^2} \\
& \delta_1(\mathbf{\hat{D}}, \mathbf{D}^{\text{gt}}) = \frac{1}{HW} \sum_{i,j}^{W,H} \mathbf{1} \left( \max \left( \frac{\mathbf{\hat{D}}_{i,j}}{\mathbf{D}^{\text{gt}}_{i,j}}, \frac{\mathbf{D}^{\text{gt}}_{i,j}}{\mathbf{\hat{D}}_{i,j}} \right) < 1.25 \right) 
\end{split}
\end{equation*}

\section{Accuracy Breakdown}
\label{sec:appendix_acc_breakdown}

More quantitative results are shown in \cref{tab:results_virtual_pattern_indoor,tab:results_virtual_pattern_outdoor,tab:results_real_pattern_appendix}. Compared to Tab.2 in the main paper, we separate the results for indoor and outdoor scenes. Our method works better than baselines under almost all settings.

\section{Qualitative Comparison}
\label{sec:qualitative_2D}

\begin{figure*}
    \centering
    \includegraphics[width=0.94\linewidth]{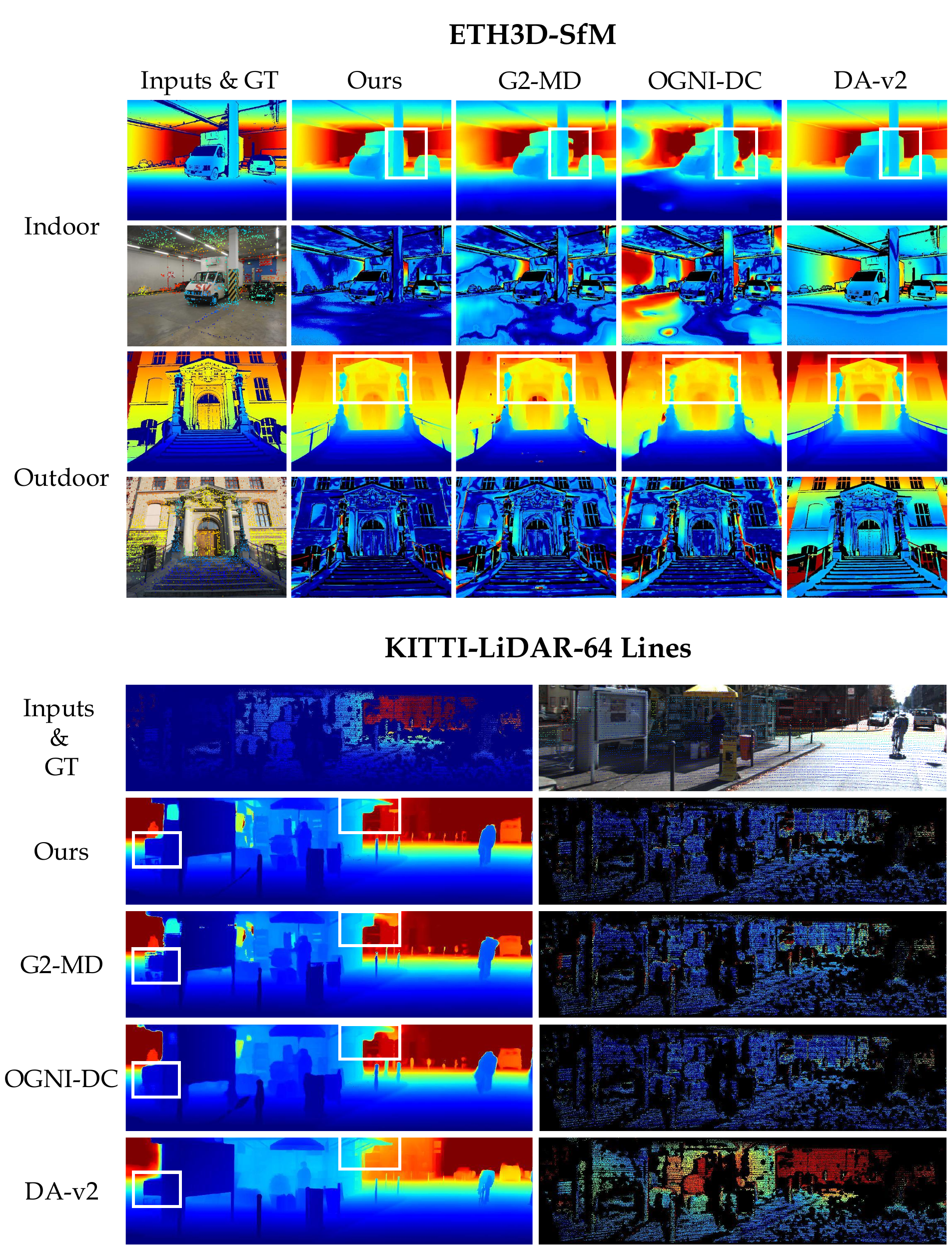}
    \caption{First row/column: gt and predicted depth; second row/column: RGB, sparse depth (superimposed), and error maps (blue means small errors).}
    \label{fig:2d_vis_p3}
\end{figure*}

\begin{figure*}
    \centering
    \vspace{-6mm}
    \includegraphics[width=0.93\linewidth]{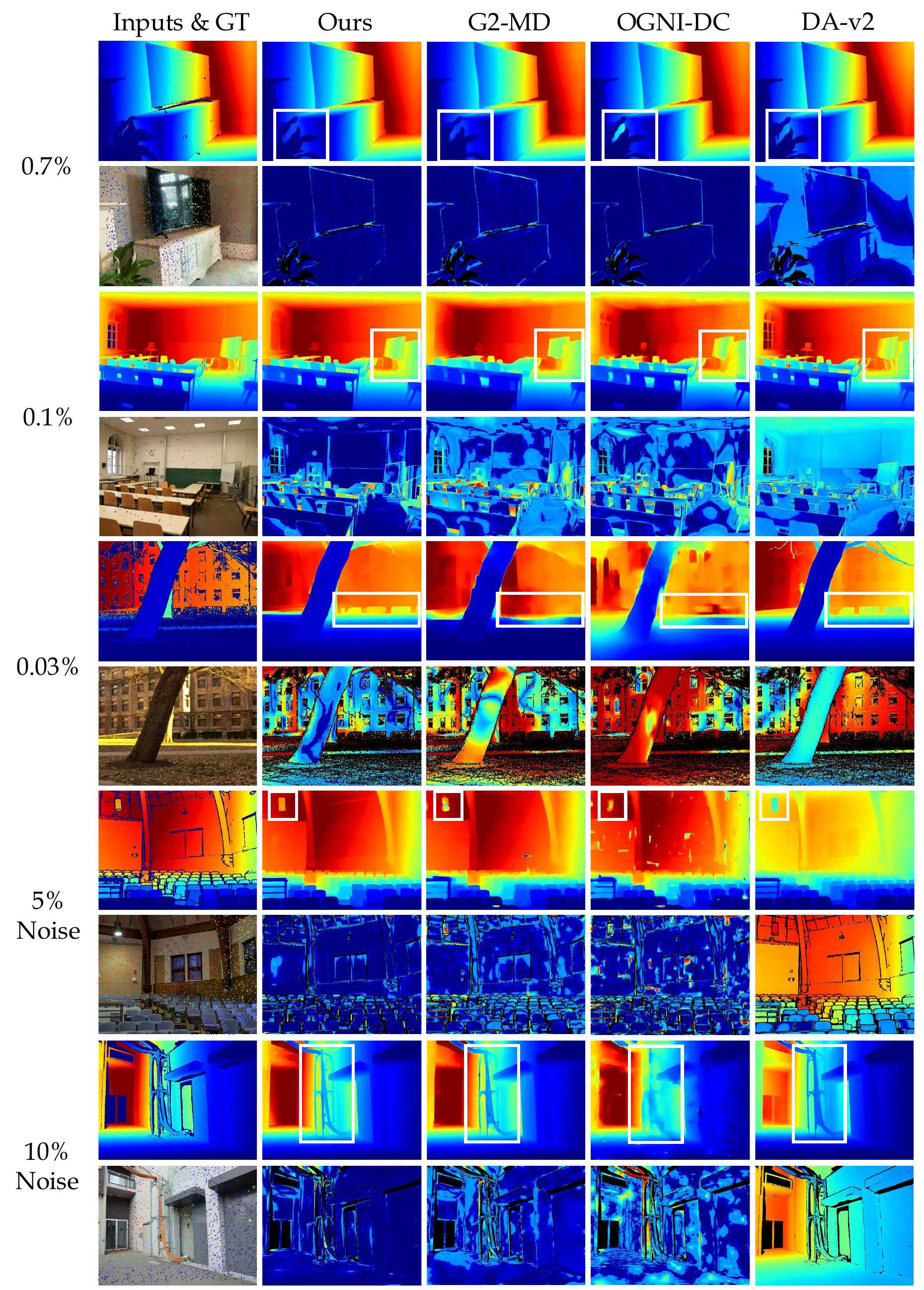}
    \caption{First row: gt and predicted depth; second row: RGB, sparse depth (superimposed), and error maps (blue means small errors).}
    \label{fig:2d_vis_p1}
\end{figure*}

\begin{figure*}
    \centering
    \vspace{-6mm}
    \includegraphics[width=0.93\linewidth]{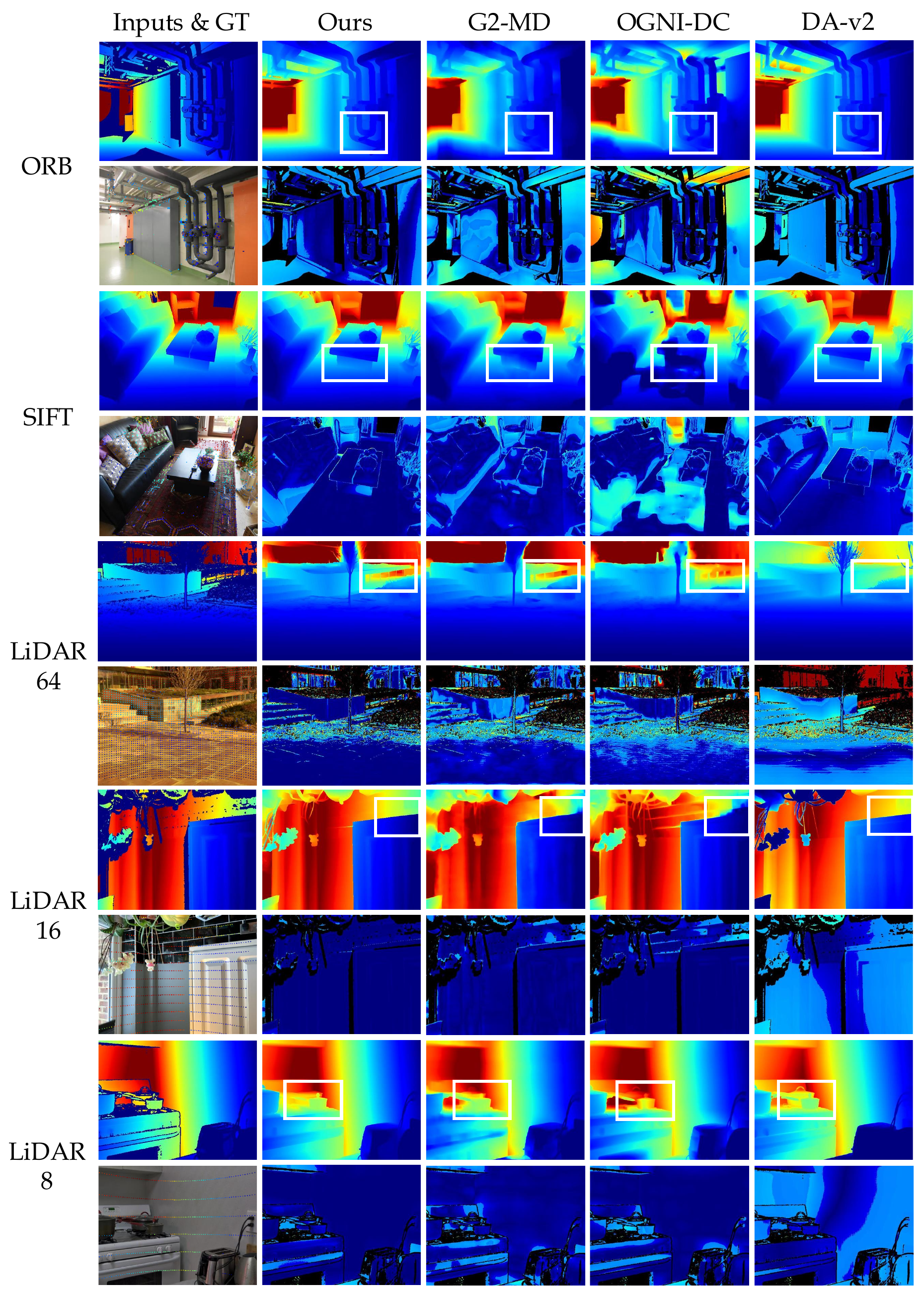}
    \caption{First row: gt and predicted depth; second row: RGB, sparse depth (superimposed), and error maps (blue means small errors).}
    \label{fig:2d_vis_p2}
\end{figure*}

Visualizations are provided in \cref{fig:2d_vis_p3,fig:2d_vis_p1,fig:2d_vis_p2}. Compared to DC methods G2-MD~\cite{g2-monodepth} and OGNI-DC~\cite{ognidc}, our method generates much sharper results and is more robust to noise. While DA-v2~\cite{depth_anything_v2} produces sharp details, its global structure is always off, especially for outdoor scenes.

\section{More Ablations on the Laplacian Loss}
\label{sec:appendix_laplacian_loss}

To show the necessity of using an $L_1$ loss along with $L_{lap}$, we conduct additional ablation studies as shown in \cref{tab:ablation_loss}. Our solution with $L_1$ works the best. This is because DC is a dense prediction task, \ie, the error on every pixel contributes to the final metrics. While $L_{lap}$ helps convergence, it falls short of enforcing a reasonable depth for every pixel. 

\begin{table}[ht]
\setlength{\tabcolsep}{1.0mm}
\footnotesize
 \vspace{-1mm}
\centering
        \caption{Ablations on removing the $L_1$ loss.}
         \vspace{-2mm}
        \label{tab:ablation_loss}
            \begin{tabular}{@{}lccccc@{}}
              \hline
   
    & 
  ETH-MAE & ETH-REL &
  KITTI-MAE & KITTI-REL \\
  
  \hline 
  $L_{lap}$ & 11.208 & 1.410 & 1.353 & 0.307  \\
  $L_{Lap}+L_{gm}$ & 0.525 & 0.081 & 1.179 & 0.282 \\
  $\bm{L_{Lap}}$+$\bm{L_{gm}}$+$\bm{L_1}$ & \textBF{0.490} & \textBF{0.076} & \textBF{1.173} & \textBF{0.277} \\
  \hline
  \end{tabular}
\end{table}

\begin{table*}[t]
\setlength{\tabcolsep}{0.13cm}
\centering
        \vspace{-2mm}
        \caption{Quantitative comparison with baselines on the \textbf{synthetic depth patterns} on the \textbf{indoor scenes}. Results averaged on the ARKitScenes, iBims, ETH3D-indoor, and DIODE-indoor subsets. }
        \vspace{-2mm}
        \label{tab:results_virtual_pattern_indoor}
            \begin{tabular}{@{}lP{1cm}P{1cm}P{1cm}P{1cm}P{1cm}P{1cm}P{1cm}P{1cm}P{1cm}P{1cm}P{1cm}P{1cm}@{}}
            
  \hline
  \noalign{\vskip 0.5mm}
  \multirow{2}{*}{Methods} &
  \multicolumn{4}{c}{0.7\%} &
  \multicolumn{4}{c}{0.1\%} &
  \multicolumn{4}{c}{0.03\%} 
  \\
  \cmidrule(lr){2-5} \cmidrule(lr){6-9} \cmidrule(lr){10-13}
   &
  RMSE & MAE & REL & $\delta_1$ &
  RMSE & MAE & REL & $\delta_1$ &
  RMSE & MAE & REL & $\delta_1$
  \\
  \hline \hline  
  \noalign{\vskip 0.5mm}
  
      Depth Pro~\cite{depth_pro} & 0.636	& 0.524	& 0.176	& 0.746	& 0.636	& 0.524	& 0.176	& 0.746	& 0.636	& 0.524	& 0.176	& 0.746 \\
      DA-v2~\cite{depth_anything_v2} & 0.626	& 0.193	& 0.042	& 0.982	& 0.632	& 0.194	& 0.042	& 0.982	& 0.636	& 0.195	& 0.042	& 0.981 \\
  Marigold~\cite{marigold} & 0.306	& 0.182	& 0.060	& 0.954	& 0.309	& 0.184	& 0.060	& 0.952	& 0.314	& 0.186	& 0.061	& 0.952 \\
  CFormer~\cite{completionformer} & 0.151	& 0.025	& 0.006	& 0.996	& 0.883	& 0.557	& 0.161	& 0.679	& 1.417	& 1.042	& 0.301	& 0.432 \\
  DFU~\cite{DFU} & 2.166	& 1.425	& 1.118	& 0.508	& 3.930	& 2.941	& 2.002	& 0.267	& 5.920	& 4.659	& 3.073	& 0.140 \\
  BP-Net~\cite{BP-Net} & 0.236	& 0.044	& 0.014	& 0.983	& 0.709	& 0.454	& 0.139	& 0.748	& 1.009	& 0.744	& 0.216	& 0.511 \\
  OGNI-DC~\cite{ognidc} & 0.105	& 0.020	& 0.005	& \textBF{0.997}	& 0.236	& 0.078	& 0.017	& 0.990	& 0.421	& 0.199	& 0.049	& 0.958 \\
  G2-MD~\cite{g2-monodepth} & 0.107	& 0.024	& 0.007	& \textBF{0.997}	& 0.195	& 0.065	& 0.019	& 0.989	& 0.327	& 0.163	& 0.056	& 0.955 \\

  \hline
  \textbf{Ours} & \textBF{0.084}	& \textBF{0.015}	& \textBF{0.004}	& \textBF{0.997}	& \textBF{0.151}	& \textBF{0.038}	& \textBF{0.010}	& \textBF{0.994}	& \textBF{0.233}	& \textBF{0.076}	& \textBF{0.020}	& \textBF{0.987} \\
  \hline
  
  \noalign{\vskip 1.5mm}

  \hline
  \noalign{\vskip 0.5mm}
  \multirow{2}{*}{Methods} &
  \multicolumn{4}{c}{5\% Noise} &
  \multicolumn{4}{c}{10 \% Noise} &
  \multicolumn{4}{c}{ORB~\cite{ORB}} 
  \\
  \cmidrule(lr){2-5} \cmidrule(lr){6-9} \cmidrule(lr){10-13}
   &
  RMSE & MAE & REL & $\delta_1$ &
  RMSE & MAE & REL & $\delta_1$ &
  RMSE & MAE & REL & $\delta_1$
  \\
  \hline \hline  
  \noalign{\vskip 0.5mm}
  
      Depth Pro~\cite{depth_pro} & 0.636	& 0.524	& 0.176	& 0.746	& 0.636	& 0.524	& 0.176	& 0.746	& 0.636	& 0.524	& 0.176	& 0.746 \\
      DA-v2~\cite{depth_anything_v2} & 1.079	& 0.527	& 0.217	& 0.857	& 1.793	& 0.851	& 0.339	& 0.701	& 1.507	& 1.123	& 0.797	& 0.963  \\
  Marigold~\cite{marigold} & 0.318	& 0.190	& 0.063	& 0.954	& 0.347	& 0.217	& 0.072	& 0.949	& 0.426	& 0.311	& 0.131	& 0.893 \\
  CFormer~\cite{completionformer} & 0.253	& 0.056	& 0.017	& 0.983	& 0.335	& 0.096	& 0.031	& 0.965	& 1.420	& 1.059	& 0.339	& 0.415 \\
  DFU~\cite{DFU} & 2.220	& 1.463	& 1.114	& 0.496	& 2.267	& 1.507	& 1.114	& 0.481	& 5.611	& 4.190	& 2.949	& 0.260 \\
  BP-Net~\cite{BP-Net} & 0.315	& 0.089	& 0.030	& 0.964	& 0.393	& 0.142	& 0.050	& 0.939	& 1.228	& 0.906	& 0.354	& 0.422 \\
  OGNI-DC~\cite{ognidc} & 0.202	& 0.047	& 0.014	& 0.986	& 0.283	& 0.084	& 0.027	& 0.970	& 0.656	& 0.438	& 0.171	& 0.713 \\
  G2-MD~\cite{g2-monodepth} & 0.134	& 0.029	& 0.008	& 0.996	& 0.155	& 0.034	& 0.009	& 0.995	& 0.438	& 0.280	& 0.124	& 0.824 \\

  \hline
  \textbf{Ours} & \textBF{0.090}	& \textBF{0.016}	& \textBF{0.004}	& \textBF{0.997}	& \textBF{0.097}	& \textBF{0.019}	& \textBF{0.005}	& \textBF{0.997}	& \textBF{0.240}	& \textBF{0.127}	& \textBF{0.057}	& \textBF{0.944}  \\
  \hline
  
  \noalign{\vskip 1.5mm}

  \hline
  \noalign{\vskip 0.5mm}
  \multirow{2}{*}{Methods} &
  \multicolumn{4}{c}{SIFT~\cite{SIFT}} &
  \multicolumn{4}{c}{LiDAR-64-Lines} &
  \multicolumn{4}{c}{LiDAR-16-Lines} 
  \\
  \cmidrule(lr){2-5} \cmidrule(lr){6-9} \cmidrule(lr){10-13}
   &
  RMSE & MAE & REL & $\delta_1$ &
  RMSE & MAE & REL & $\delta_1$ &
  RMSE & MAE & REL & $\delta_1$
  \\
  \hline \hline  
  \noalign{\vskip 0.5mm}
  
      Depth Pro~\cite{depth_pro} & 0.636	& 0.524	& 0.176	& 0.746	& 0.636	& 0.524	& 0.176	& 0.746	& 0.636	& 0.524	& 0.176	& 0.746 \\
      DA-v2~\cite{depth_anything_v2} & 0.749	& 0.549	& 0.390	& \textBF{0.973}	& 2.359	& 0.300	& 0.108	& 0.980	& 0.597	& 0.189	& 0.041	& 0.982 \\
  Marigold~\cite{marigold} & 0.413	& 0.301	& 0.127	& 0.905	& 1.166	& 0.182	& 0.060	& 0.954	& 0.306	& 0.182	& 0.060	& 0.954 \\
  CFormer~\cite{completionformer} & 1.315	& 0.978	& 0.317	& 0.442	& 3.473	& 0.017	& \textBF{0.004}	& \textBF{0.997}	& 0.255	& 0.075	& 0.020	& 0.981 \\
  DFU~\cite{DFU} & 5.721	& 4.305	& 2.992	& 0.239	& 5.277	& 1.472	& 1.319	& 0.629	& 2.455	& 1.726	& 1.361	& 0.449\\
  BP-Net~\cite{BP-Net} & 1.150	& 0.836	& 0.328	& 0.469	& 2.217	& 0.037	& 0.012	& 0.985	& 0.346	& 0.110	& 0.036	& 0.954 \\
  OGNI-DC~\cite{ognidc} & 0.517	& 0.332	& 0.134	& 0.807	& 1.242	& \textBF{0.016}	& \textBF{0.004}	& \textBF{0.997}	& 0.154	& 0.040	& 0.009	& 0.995 \\
  G2-MD~\cite{g2-monodepth} & 0.402	& 0.257	& 0.117	& 0.834	& 0.882	& 0.022	& 0.006	& \textBF{0.997}	& 0.150	& 0.045	& 0.012	& 0.994 \\

  \hline
  \textbf{Ours} & \textBF{0.203}	& \textBF{0.101}	& \textBF{0.046}	& 0.960	& \textBF{0.611}	& \textBF{0.016}	& \textBF{0.004}	& \textBF{0.997}	& \textBF{0.107}	& \textBF{0.024}	& \textBF{0.006}	& \textBF{0.996} \\
  \hline
  
  \noalign{\vskip 1.5mm}

  \hhline{-----}
  \noalign{\vskip 0.5mm}
  \multirow{2}{*}{Methods} &
  \multicolumn{4}{c}{LiDAR-8-Lines} &
  \\
  \cmidrule(lr){2-5}
   &
  RMSE & MAE & REL & $\delta_1$ &
  \\
  \hhline{=====}
  \noalign{\vskip 0.5mm}
  
      Depth Pro~\cite{depth_pro} & 0.636	& 0.524	& 0.176	& 0.746 \\
      DA-v2~\cite{depth_anything_v2} & 0.602	& 0.194	& 0.042	& 0.982 \\
  Marigold~\cite{marigold} & 0.309	& 0.187	& 0.062	& 0.951 \\
  CFormer~\cite{completionformer} & 0.934	& 0.609	& 0.168	& 0.662 \\
  DFU~\cite{DFU} & 4.022	& 3.029	& 2.141	& 0.257 \\
  BP-Net~\cite{BP-Net} & 0.816	& 0.587	& 0.179	& 0.652 \\
  OGNI-DC~\cite{ognidc} & 0.287	& 0.114	& 0.028	& 0.979 \\
  G2-MD~\cite{g2-monodepth} & 0.219	& 0.083	& 0.023	& 0.988 \\

  \hhline{-----}
  \textbf{Ours} & \textBF{0.163}	& \textBF{0.050}	& \textBF{0.014}	& \textBF{0.993} \\
  \hhline{-----}
  \end{tabular}
\end{table*}

\begin{table*}[t]
\setlength{\tabcolsep}{0.13cm}
\centering
        \vspace{-2mm}
        \caption{Quantitative comparison with baselines on the \textbf{synthetic depth patterns} on the \textbf{outdoor scenes}. Results averaged on the ETH3D-outdoor and DIODE-outdoor subsets. }
        \vspace{-2mm}
        \label{tab:results_virtual_pattern_outdoor}
            \begin{tabular}{@{}lP{1cm}P{1cm}P{1cm}P{1cm}P{1cm}P{1cm}P{1cm}P{1cm}P{1cm}P{1cm}P{1cm}P{1cm}@{}}
            
  \hline
  \noalign{\vskip 0.5mm}
  \multirow{2}{*}{Methods} &
  \multicolumn{4}{c}{0.7\%} &
  \multicolumn{4}{c}{0.1\%} &
  \multicolumn{4}{c}{0.03\%} 
  \\
  \cmidrule(lr){2-5} \cmidrule(lr){6-9} \cmidrule(lr){10-13}
   &
  RMSE & MAE & REL & $\delta_1$ &
  RMSE & MAE & REL & $\delta_1$ &
  RMSE & MAE & REL & $\delta_1$
  \\
  \hline \hline  
  \noalign{\vskip 0.5mm}
  
      Depth Pro~\cite{depth_pro} & 7.712	& 6.368	& 0.426	& 0.183	& 7.712	& 6.368	& 0.426	& 0.183	& 7.712	& 6.368	& 0.426	& 0.183 \\
      DA-v2~\cite{depth_anything_v2} & 6.003	& 1.993	& 0.114	& 0.924	& 6.195	& 2.103	& 0.116	& 0.919	& 6.314	& 2.118	& 0.121	& 0.922  \\
  Marigold~\cite{marigold} & 2.454	& 1.351	& 0.123	& 0.884	& 2.514	& 1.382	& 0.124	& 0.882	& 2.619	& 1.425	& 0.130	& 0.881 \\
  CFormer~\cite{completionformer} & 4.999	& 3.239	& 0.663	& 0.625	& 9.578	& 7.504	& 1.437	& 0.360	& 12.149	& 10.198	& 1.875	& 0.240 \\
  DFU~\cite{DFU} & 2.771	& 1.255	& 0.158	& 0.850	& 5.486	& 3.198	& 0.440	& 0.609	& 7.504	& 4.779	& 0.685	& 0.466 \\
  BP-Net~\cite{BP-Net} & 3.046	& 1.281	& 0.102	& 0.917	& 6.368	& 3.766	& 0.276	& 0.758	& 7.112	& 4.379	& 0.340	& 0.672 \\
  OGNI-DC~\cite{ognidc} &  1.747	& 0.554	& 0.046	& 0.967	& 2.974	& 1.449	& 0.169	& 0.855	& 4.140	& 2.484	& 0.330	& 0.710 \\
  G2-MD~\cite{g2-monodepth} & 1.453	& 0.368	& 0.032	& 0.980	& 2.261	& 0.868	& 0.086	& 0.933	& 3.235	& 1.772	& 0.171	& 0.803 \\

  \hline
  \textbf{Ours} & \textBF{1.275}	& \textBF{0.292}	& \textBF{0.022}	& \textBF{0.985}	& \textBF{1.889}	& \textBF{0.599}	& \textBF{0.044}	& \textBF{0.967}	& \textBF{2.477}	& \textBF{0.970}	& \textBF{0.070}	& \textBF{0.942} \\
  \hline
  
  \noalign{\vskip 1.5mm}

  \hline
  \noalign{\vskip 0.5mm}
  \multirow{2}{*}{Methods} &
  \multicolumn{4}{c}{5\% Noise} &
  \multicolumn{4}{c}{10 \% Noise} &
  \multicolumn{4}{c}{ORB~\cite{ORB}} 
  \\
  \cmidrule(lr){2-5} \cmidrule(lr){6-9} \cmidrule(lr){10-13}
   &
  RMSE & MAE & REL & $\delta_1$ &
  RMSE & MAE & REL & $\delta_1$ &
  RMSE & MAE & REL & $\delta_1$
  \\
  \hline \hline  
  \noalign{\vskip 0.5mm}
  
      Depth Pro~\cite{depth_pro} & 7.712	& 6.368	& 0.426	& 0.183	& 7.712	& 6.368	& 0.426	& 0.183	& 7.712	& 6.368	& 0.426	& 0.183 \\
      DA-v2~\cite{depth_anything_v2} & 8.689	& 4.452	& 0.281	& 0.646	& 10.893	& 6.302	& 0.463	& 0.350	& 5.066	& 2.026	& 0.112	& 0.895 \\
  Marigold~\cite{marigold} & 2.505	& 1.390	& 0.123	& 0.887	& 2.630	& 1.512	& 0.129	& 0.882	& 2.738	& 1.637	& 0.156	& 0.825 \\
  CFormer~\cite{completionformer} & 5.064	& 3.316	& 0.674	& 0.617	& 5.133	& 3.401	& 0.686	& 0.608	& 7.577	& 4.988	& 0.979	& 0.544 \\
  DFU~\cite{DFU} & 3.262	& 1.620	& 0.185	& 0.800	& 3.713	& 1.995	& 0.213	& 0.747	& 4.376	& 2.469	& 0.370	& 0.655 \\
  BP-Net~\cite{BP-Net} & 3.120	& 1.340	& 0.113	& 0.901	& 3.242	& 1.441	& 0.129	& 0.879	& 4.302	& 2.112	& 0.205	& 0.805 \\
  OGNI-DC~\cite{ognidc} & 1.962	& 0.690	& 0.057	& 0.954	& 2.160	& 0.822	& 0.069	& 0.940	& 3.019	& 1.480	& 0.194	& 0.826 \\
  G2-MD~\cite{g2-monodepth} & 1.553	& 0.402	& 0.034	& 0.978	& 1.663	& 0.442	& 0.035	& 0.975	& 2.019	& 0.794	& 0.081	& 0.920 \\

  \hline
  \textbf{Ours} & \textBF{1.323}	& \textBF{0.313}	& \textBF{0.023}	& \textBF{0.983}	& \textBF{1.390}	& \textBF{0.341}	& \textBF{0.024}	& \textBF{0.982}	& \textBF{1.646}	& \textBF{0.514}	& \textBF{0.039}	& \textBF{0.967} \\
  \hline
  
  \noalign{\vskip 1.5mm}

  \hline
  \noalign{\vskip 0.5mm}
  \multirow{2}{*}{Methods} &
  \multicolumn{4}{c}{SIFT~\cite{SIFT}} &
  \multicolumn{4}{c}{LiDAR-64-Lines} &
  \multicolumn{4}{c}{LiDAR-16-Lines} 
  \\
  \cmidrule(lr){2-5} \cmidrule(lr){6-9} \cmidrule(lr){10-13}
   &
  RMSE & MAE & REL & $\delta_1$ &
  RMSE & MAE & REL & $\delta_1$ &
  RMSE & MAE & REL & $\delta_1$
  \\
  \hline \hline  
  \noalign{\vskip 0.5mm}
  
      Depth Pro~\cite{depth_pro} & 7.712	& 6.368	& 0.426	& 0.183	& 7.712	& 6.368	& 0.426	& 0.183	& 7.712	& 6.368	& 0.426	& 0.183 \\
      DA-v2~\cite{depth_anything_v2} & 5.580	& 2.082	& 0.116	& 0.905	& 5.918	& 1.960	& 0.113	& 0.924	& 6.033	& 2.030	& 0.114	& 0.923 \\
  Marigold~\cite{marigold} & 2.671	& 1.583	& 0.155	& 0.847	& 2.451	& 1.340	& 0.123	& 0.884	& 2.468	& 1.349	& 0.124	& 0.883 \\
  CFormer~\cite{completionformer} & 7.788	& 5.450	& 1.125	& 0.507	& 3.351	& 1.758	& 0.339	& 0.771	& 4.424	& 2.628	& 0.513	& 0.696 \\
  DFU~\cite{DFU} & 4.388	& 2.475	& 0.408	& 0.662	& 2.975	& 1.191	& 0.181	& 0.844	& 3.380	& 1.656	& 0.192	& 0.815 \\
  BP-Net~\cite{BP-Net} & 4.352	& 2.174	& 0.239	& 0.807	& 2.234	& 0.787	& 0.075	& 0.937	& 4.505	& 2.243	& 0.160	& 0.873 \\
  OGNI-DC~\cite{ognidc} & 2.690	& 1.299	& 0.185	& 0.837	& 1.550	& 0.435	& 0.035	& 0.974	& 2.157	& 0.831	& 0.081	& 0.937 \\
  G2-MD~\cite{g2-monodepth} & 1.844	& 0.677	& 0.077	& 0.925	& \textBF{1.200}	& \textBF{0.292}	& 0.025	& \textBF{0.985}	& 1.756	& 0.524	& 0.047	& 0.970 \\

  \hline
  \textbf{Ours} & \textBF{1.429}	& \textBF{0.403}	& \textBF{0.034}	& \textBF{0.974}	& 1.271	& 0.303	& \textBF{0.023}	& 0.983	& \textBF{1.513}	& \textBF{0.412}	& \textBF{0.031}	& \textBF{0.978} \\
  \hline
  
  \noalign{\vskip 1.5mm}

  \hhline{-----}
  \noalign{\vskip 0.5mm}
  \multirow{2}{*}{Methods} &
  \multicolumn{4}{c}{LiDAR-8-Lines} &
  \\
  \cmidrule(lr){2-5}
   &
  RMSE & MAE & REL & $\delta_1$ &
  \\
  \hhline{=====}
  \noalign{\vskip 0.5mm}
  
      Depth Pro~\cite{depth_pro} & 7.712	& 6.368	& 0.426	& 0.183	\\
      DA-v2~\cite{depth_anything_v2} & 6.304	& 2.056	& 0.119	& 0.922 \\
  Marigold~\cite{marigold} &  2.578	& 1.382	& 0.124	& 0.883 \\
  CFormer~\cite{completionformer} &  7.759	& 5.549	& 1.071	& 0.472 \\
  DFU~\cite{DFU} & 5.242	& 3.027	& 0.401	& 0.623 \\
  BP-Net~\cite{BP-Net} & 5.859	& 3.282	& 0.226	& 0.776 \\
  OGNI-DC~\cite{ognidc} & 3.354	& 1.671	& 0.197	& 0.824 \\
  G2-MD~\cite{g2-monodepth} & 2.404	& 0.918	& 0.078	& 0.936\\

  \hhline{-----}
  \textbf{Ours} & \textBF{2.096}	& \textBF{0.715}	& \textBF{0.048}	& \textBF{0.961} \\
  \hhline{-----} 
  \end{tabular}
\end{table*}

\begin{table*}[t]
\setlength{\tabcolsep}{0.13cm}
\centering
        \vspace{-2mm}
        \caption{Quantitative comparison with baselines on the ETH3D-SfM and KITTIDC. The numbers in \textcolor{gray}{gray} are trained on KITTI and excluded from the ranking.}
        \vspace{-2mm}
        \label{tab:results_real_pattern_appendix}
            \begin{tabular}{@{}lP{1cm}P{1cm}P{1cm}P{1cm}P{1cm}P{1cm}P{1cm}P{1cm}P{1cm}P{1cm}P{1cm}P{1cm}@{}}
            
  \hline
  \noalign{\vskip 0.5mm}
  \multirow{2}{*}{Methods} &
  \multicolumn{4}{c}{ETH3D-SfM-Indoor} &
  \multicolumn{4}{c}{ETH3D-SfM-Outdoor} &
  \multicolumn{4}{c}{KITTI-64-Lines} 
  \\
  \cmidrule(lr){2-5} \cmidrule(lr){6-9} \cmidrule(lr){10-13}
   &
  RMSE & MAE & REL & $\delta_1$ &
  RMSE & MAE & REL & $\delta_1$ &
  RMSE & MAE & REL & $\delta_1$
  \\
  \hline \hline  
  \noalign{\vskip 0.5mm}

  CFormer~\cite{completionformer} & 2.088	& 0.811	& 0.229	& 0.616	& 9.108	& 4.782	& 1.215	& 0.520	& 
  \textcolor{gray}{0.741}	& \textcolor{gray}{0.195}	&\textcolor{gray}{0.011} &	\textcolor{gray}{0.998} \\
  DFU~\cite{DFU} & 3.572	& 2.417	& 1.105	& 0.446	& 4.296	& 2.494	& 0.588	& 0.624	& \textcolor{gray}{0.713} &	\textcolor{gray}{0.186}	& \textcolor{gray}{0.010} & \textcolor{gray}{0.998} \\
  BP-Net~\cite{BP-Net} & 1.664	& 0.864	& 0.301	& 0.600	& 4.342	& 1.859	& 0.339	& 0.770	& \textcolor{gray}{0.784} &	\textcolor{gray}{0.204} & \textcolor{gray}{0.011} & \textcolor{gray}{0.998} \\
  DPromting~\cite{depth_prompting} & 1.306 & 1.004 & 0.269 & 0.605 & 5.596 & 4.664 & 0.846 & 0.349 & \textcolor{gray}{1.078} & \textcolor{gray}{0.324} & \textcolor{gray}{0.019} & \textcolor{gray}{0.993} \\
  OGNI-DC~\cite{ognidc} & 1.108	& 0.520	& 0.181	& 0.758	& 2.671	& 1.270	& 0.268	& 0.787	& \textcolor{gray}{0.750} &	\textcolor{gray}{0.193} & \textcolor{gray}{0.010} & \textcolor{gray}{0.998} \\
  \hhline{>{\arrayrulecolor [gray]{1.0}}--------->{\arrayrulecolor {black}}----}
  
      Depth Pro~\cite{depth_pro} & 0.928	& 0.749	& 0.208	& 0.659	& 5.433	& 4.824	& 0.441	& 0.196	& 4.893	& 3.233 & 0.211 & 0.651	 \\
      DA-v2~\cite{depth_anything_v2} & \textBF{0.592}	& 0.280	& \textBF{0.065}	& \textBF{0.950}	& 2.663	& 0.805	& 0.082	& 0.935	& 4.561 & 1.925 & 0.090 & 0.924 \\
  Marigold~\cite{marigold} & 0.627	& 0.472	& 0.152	& 0.842	& 1.883	& 1.270	& 0.252	& 0.715	& 3.462	& 1.911 & 0.118 & 0.889 \\
  
  G2-MD~\cite{g2-monodepth} & 1.068	& 0.416	& 0.164	& 0.896	& 2.453	& 0.770	& 0.153	& 0.889	& 1.612	& 0.376 & 0.024 & 0.986 \\

  \hline

  \textbf{Ours} & 0.605	& \textBF{0.239}	& 0.090	& 0.932	& \textBF{1.069}	& \textBF{0.312}	& \textBF{0.053}	& \textBF{0.953}	& \textBF{1.191} &	\textBF{0.270} &	\textBF{0.015} &	\textBF{0.993} \\
  \hline
  
  \noalign{\vskip 1.5mm}

  \hline
  \noalign{\vskip 0.5mm}
  \multirow{2}{*}{Methods} &
  \multicolumn{4}{c}{KITTI-32-Lines} &
  \multicolumn{4}{c}{KITTI-16-Lines} &
  \multicolumn{4}{c}{KITTI-8-Lines} 
  \\
  \cmidrule(lr){2-5} \cmidrule(lr){6-9} \cmidrule(lr){10-13}
   &
  RMSE & MAE & REL & $\delta_1$ &
  RMSE & MAE & REL & $\delta_1$ &
  RMSE & MAE & REL & $\delta_1$
  \\
  \hline \hline  
  \noalign{\vskip 0.5mm}

  CFormer~\cite{completionformer} & 
   \textcolor{gray}{1.245}	& \textcolor{gray}{0.387}	& 	 \textcolor{gray}{0.022}	& 	 \textcolor{gray}{0.991}	& 	 \textcolor{gray}{2.239}	& 	 \textcolor{gray}{0.882}	& 	 \textcolor{gray}{0.050}	& 	 \textcolor{gray}{0.969}	& 	 \textcolor{gray}{3.650}	& 	 \textcolor{gray}{1.701}	& 	 \textcolor{gray}{0.102}	& 	 \textcolor{gray}{0.877}
 \\
  DFU~\cite{DFU} & 
  \textcolor{gray}{1.099}	&	\textcolor{gray}{0.315}	&	\textcolor{gray}{0.018}	&	\textcolor{gray}{0.995}	&	\textcolor{gray}{2.070}	&	\textcolor{gray}{0.738}	&	\textcolor{gray}{0.040}	&	\textcolor{gray}{0.976}	&	\textcolor{gray}{3.269}	&	\textcolor{gray}{1.468}	&	\textcolor{gray}{0.08}	&	\textcolor{gray}{0.915} \\
  BP-Net~\cite{BP-Net} & 
  \textcolor{gray}{1.032}	&   \textcolor{gray}{0.296}	&	\textcolor{gray}{0.016}	&	\textcolor{gray}{0.996}	&	\textcolor{gray}{1.524}	&	\textcolor{gray}{0.490}	&	\textcolor{gray}{0.026}	&	\textcolor{gray}{0.991}	&	\textcolor{gray}{2.391}	&	\textcolor{gray}{0.953}	&	\textcolor{gray}{0.052} & \textcolor{gray}{0.971} \\
  DPromting~\cite{depth_prompting} & \textcolor{gray}{1.234} & \textcolor{gray}{0.382} & \textcolor{gray}{0.021} & \textcolor{gray}{0.992} & \textcolor{gray}{1.475} & \textcolor{gray}{0.477} & \textcolor{gray}{0.025} & \textcolor{gray}{0.990} & \textcolor{gray}{1.7907} & \textcolor{gray}{0.6344} & \textcolor{gray}{0.0322} & \textcolor{gray}{0.986} \\
  OGNI-DC~\cite{ognidc} & \textcolor{gray}{1.018} & \textcolor{gray}{0.268} & \textcolor{gray}{0.014} & \textcolor{gray}{0.996} & \textcolor{gray}{1.664} & \textcolor{gray}{0.453} & \textcolor{gray}{0.022} & \textcolor{gray}{0.990}	& \textcolor{gray}{2.363} & \textcolor{gray}{0.777} & \textcolor{gray}{0.039} &	\textcolor{gray}{0.977} \\
  \hline
  
      Depth Pro~\cite{depth_pro} & 4.893	& 3.233 & 0.211 & 0.651 & 4.893	& 3.233 & 0.211 & 0.651 & 4.893	& 3.233 & 0.211 & 0.651 \\
      DA-v2~\cite{depth_anything_v2} & 4.583 & 1.928 & 0.090 & 0.923 & 4.615 & 1.934 & 0.090 & 0.923 & 4.689 & 1.951 & 0.091 & 0.922 \\
  Marigold~\cite{marigold} & 3.463 & 1.902 & 0.117 & 0.892 & 3.468 & 1.904 & 0.117 & 0.891 & 3.498 & 1.939 & 0.120 & 0.885 \\
  
  G2-MD~\cite{g2-monodepth} & 1.802 & 0.447 & 0.027 & 0.985 & 2.222 & 0.645 & 0.035 & 0.981 & 2.769 & 0.901 & 0.046 & 0.970 \\

 \hline
  \textbf{Ours} & \textBF{1.398} &	\textBF{0.339} & \textBF{0.019} &	\textBF{0.990} & \textBF{1.682} & \textBF{0.441} & \textBF{0.023} & \textBF{0.987} & \textBF{2.058} & \textBF{0.597} & \textBF{0.030} & \textBF{0.982} \\
  \hline
  \end{tabular}
\end{table*}

\clearpage

\end{document}